\let\TeXyear\year
\documentclass{IEEEaccess}

\let\year\TeXyear
\usepackage{amsmath,amssymb,amsfonts}
\usepackage{graphicx}
\usepackage{textcomp}

\usepackage{xcolor}
\usepackage{hyperref}
\usepackage{siunitx}
\usepackage{physics}
\usepackage[T1]{fontenc}
\AtBeginDocument{\RenewCommandCopy\qty\SI}
\usepackage{array}
\usepackage{stfloats}
\usepackage{changepage}
\usepackage{float}
\usepackage{url}
\usepackage{verbatim}
\usepackage{subfig}
\usepackage{pifont}
\usepackage{algorithm}
\usepackage{overpic}
\usepackage{comment}
\usepackage{multirow}
\usepackage{algpseudocode}
\usepackage{booktabs}
\usepackage{rotating}
\usepackage{makecell}
\usepackage{tabularx}
\usepackage{blindtext}  
\usepackage[capitalize]{cleveref}

\usepackage{tikz}

\usepackage{pgfplots} 
\usepgfplotslibrary{groupplots}
\usepgfplotslibrary{external} 
\usetikzlibrary{arrows, arrows.meta, external} 

\pgfplotsset{compat=1.18} 
\usepackage{transparent} 

\usepackage[style=ieee, backend=biber]{biblatex}
\def\BibTeX{{\rm B\kern-.05em{\sc i\kern-.025em b}\kern-.08em
    T\kern-.1667em\lower.7ex\hbox{E}\kern-.125emX}}

\usepackage{orcidlink}
\definecolor{accessblue}{cmyk}{1, 0.3, 0, 0.2}
\definecolor{greycolor}{cmyk}{0,0,0,.8}
\definecolor{Bluelight}{HTML}{0065BD} 
\definecolor{Black}{HTML}{000000}
\definecolor{Blue}{HTML}{005293}
\definecolor{Bluestrong}{HTML}{003359}
\definecolor{Red}{HTML}{8C000F}
\definecolor{Grey}{HTML}{808080}
\definecolor{Greylight}{HTML}{CCCCCC}
\definecolor{Orange}{HTML}{E37222}
\definecolor{Green}{HTML}{A2AD00}
\definecolor{GreenCR}{HTML}{008000}
\definecolor{OrangeCR}{HTML}{f1b514}

\newcommand{\arrowCR}{
  \tikz{
    \draw[{Circle[GreenCR,length=4pt]}-{triangle 45 [GreenCR,length=4pt]}, GreenCR] (0,0) -- (0.5,0);
    \draw[line width=2pt, GreenCR] (0.05,0) -- (0.3,0);
  }
}

\newcommand{\boxCR}{
  \tikz{
    \draw[fill=OrangeCR] (0,0.0) rectangle (0.4,0.15);
  }
}

\newcommand{\egoCRsmall}{
  \tikz{
    \node[inner sep=0pt] at (0,0) {\includegraphics[height=2.1mm]{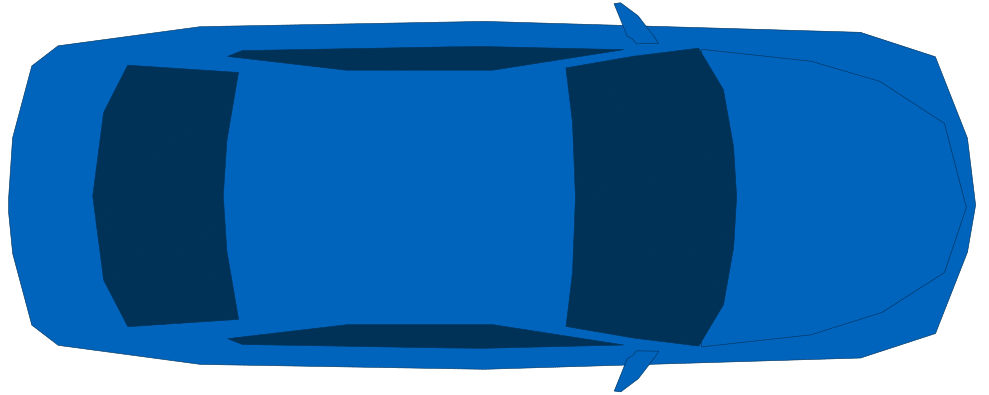}};
  }
}

\newcommand{\refpathCR}{
  \tikz{
    \draw[thick, GreenCR] (0,0) -- (0.4,0);
    \node[inner sep=0pt] at (0,-0.05) {};
  }
}

\newcommand{\trajectoriesCR}{
  \tikz{
    \draw[thick, Grey] (0,0) -- (0.4,0);
    \node[inner sep=0pt] at (0,-0.05) {};
  }
}

\newcommand{\optimaltrajectoryCR}{
  \tikz{
    \draw[thick, Bluelight] (0,0) -- (0.4,0);
    \node[inner sep=0pt] at (0,-0.05) {};
  }
}

\usepackage{fancyhdr}

\begin{document}
\history{Date of publication xxxx 00, 0000, date of current version xxxx 00, 0000.}
\doi{10.1109/ACCESS.2022.0092316              }

\title{FRENETIX: A High-Performance and Modular Motion Planning Framework for Autonomous Driving}

\author{\uppercase{Rainer Trauth}$^{\orcidlink{0000-0002-3801-6855}}$\authorrefmark{1,*},
\uppercase{Korbinian Moller}$^{\orcidlink{0000-0001-7120-0796}}$\authorrefmark{2,*},
\uppercase{Gerald Würsching}$^{\orcidlink{0000-0002-4979-6180}}$\authorrefmark{3},
and \\ \uppercase{Johannes Betz}$^{\orcidlink{0000-0001-9197-2849}}$\authorrefmark{2}, \IEEEmembership{Member, IEEE}}

\address[1]{The author is with the Institute of Automotive Technology, Technical University of Munich, 85748 Garching, Germany; Munich Institute of Robotics and Machine Intelligence (MIRMI).}
\address[2]{K. Moller, J. Betz are with the Professorship of Autonomous Vehicle Systems, TUM School of Engineering and Design, Technical University of Munich, 85748 Garching, Germany; Munich Institute of Robotics and Machine Intelligence (MIRMI).}
\address[3]{The author is with the Professorship of Cyber-physical Systems, Technical University of Munich, 85748 Garching, Germany.}
\address[*]{Shared first authorship.}

\tfootnote{The authors gratefully acknowledge the financial support from the Bavarian Research Foundation and the company Tier IV.}


\corresp{Corresponding author: Rainer Trauth (e-mail: rainer.trauth@tum.de).}

\begin{abstract}

Our research introduces a modular motion planning framework for autonomous vehicles using a sampling-based trajectory planning algorithm. This approach effectively tackles the challenges of solution space construction and optimization in path planning. The algorithm is applicable to both real vehicles and simulations, offering a robust solution for complex autonomous navigation.
Our method employs a multi-objective optimization strategy for efficient navigation in static and highly dynamic environments, focusing on optimizing trajectory comfort, safety, and path precision.
The algorithm is used to analyze the algorithm performance and success rate in 1750 virtual complex urban and highway scenarios. Our results demonstrate fast calculation times (8ms for 800 trajectories), a high success rate in complex scenarios (88\%), and easy adaptability with different modules presented. The most noticeable difference exhibited was the fast trajectory sampling, feasibility check, and cost evaluation step across various trajectory counts. 
We demonstrate the integration and execution of the framework on real vehicles by evaluating deviations from the controller using a test track. This evaluation highlights the algorithm's robustness and reliability, ensuring it meets the stringent requirements of real-world autonomous driving scenarios.
The code and the additional modules used in this research are publicly available as open-source software and can be accessed at the following link:~\url{https://github.com/TUM-AVS/Frenetix-Motion-Planner}.

\end{abstract}


\titlepgskip=-15pt

\maketitle



\section{Introduction}
\label{sec:introduction}
\begin{figure}[ht!]
    \centering
    \hspace{1mm}
    \begin{tikzpicture}[font=\scriptsize]
        \node[inner sep=0pt] at (0.7,0) {\includegraphics[height=2.5mm]{figures/ego.png}};
        \node[align=left, anchor=west] at (1.0,0) {ego \\ vehicle};
    
        \node[inner sep=0pt] at (2.6,0) {\includegraphics[height=2.5mm]{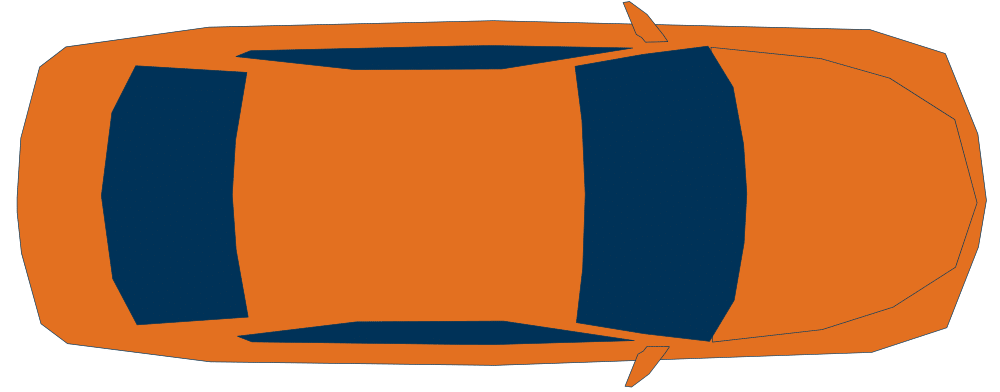}};
        \node[align=left, anchor=west] at (2.9,0) {dynamic \\ obstacle};
    
        \draw[thick, GreenCR] (4.2,0) -- (4.6,0);
        \node[align=left, anchor=west] at (4.6,0) {reference \\ path $\Gamma$};

        \node[inner sep=0pt] at (6.1,0) {\optimaltrajectoryCR};
        \node[align=left, anchor=west] at (6.3,0) {optimal \\ trajectory};

        \node[inner sep=0pt] at (7.7,0) {\trajectoriesCR};
        \node[align=left, anchor=west] at (7.9,0) {trajectory \\ samples};
        
    \end{tikzpicture}
    {\setlength{\fboxsep}{2pt}%
    \fbox{\input{figures/SamplingIntroduction}}
    }
    \caption{Visualization of a trajectory planner for AVs, depicting potential trajectories while selecting an optimal trajectory for safe navigation in a dynamic and complex urban environment.}
    \label{fig:introduction}
\end{figure}

With its promise of revolutionizing transportation, autonomous driving technology faces significant real-world challenges, as highlighted by various collision reports and practical experiences~\cite{Pokorny2022}. Among these challenges are the complexities of urban navigation, the unpredictability of traffic and pedestrian behavior, and the necessity for rapid, informed decision-making in constantly changing environments~\cite{Gu2013}. These factors underscore the importance of high-performance and adaptable trajectory planning algorithms in autonomous vehicles (AVs) (\cref{fig:introduction}).
Unfortunately, foundational, traditional trajectory planning methods often struggle to cope with the dynamic and intricate nature of real-world driving scenarios, especially in diverse and unpredictable conditions encountered in urban settings. This highlights the need for trajectory planning algorithms with low calculation times, robustness, and high adaptability to various situations to ensure safety and efficiency in autonomous driving. Our work introduces a modular framework for motion planning, including an analytical sampling-based trajectory planning algorithm, to address these challenges. This algorithm is designed to efficiently handle the complexities and uncertainties inherent in urban driving environments. In summary, this work has three main contributions:
\begin{itemize}
    \item We introduce FRENETIX, a publicly available sampling-based trajectory planner for autonomous vehicles (AVs). This planner utilizes a multi-objective optimization strategy to enhance trajectory comfort, safety, and path precision in complex environments. We now offer a comprehensive toolbox that combines a planner and simulation environment, providing an all-in-one solution for diverse scenarios.
    \item Our approach is modular, enhancing adaptability and scalability. This design allows for straightforward integration and optimal functionality of system components across various scenarios.
    \item We offer Python and C++ implementations to demonstrate the algorithm's real-time capability and success rate in complex scenarios, including tests on actual vehicles to highlight its practical effectiveness and efficiency in real-world applications.
\end{itemize}


\section{Related Work}
\label{sec:relatedwork}

Motion planning is a critical component in the software development of autonomous vehicles~\cite{Gonzlez2015}. Motion planners operate on a specific time horizon and attempt to avoid obstacles while quickly generating an efficient and reliable local trajectory. The trajectory includes a path of $x(t)$ and $y(t)$ coordinates and a velocity profile $v(t)$, which provide the reference information for subsequent control. This task is complex due to dynamic and unpredictable road environments, especially in urban settings with static and dynamic obstacles like pedestrians and other vehicles~\cite{Katrakazas2015}. Several concepts have been developed from various theoretical foundations in recent decades~\cite{Teng2023, Zhou2022}. 
The methods can be categorized into optimization-, graph-, sampling-, and machine learning-based algorithms~\cite{Paden.2016,Teng2023,Zhou2022,Gonzlez2015,Dong2023,Tampuu2022,Gonzlez2015}.
Optimization methods~\cite{Li2023, Li2022, Ziegler20214}, like optimal control problems, are known for their mathematical rigor and ability to optimize continuous paths, often leveraging calculus of variations to find smooth trajectories in complex spaces. These methods focus on finding the best solution or path under a given set of constraints and use mathematical optimization techniques to minimize or maximize an objective function. Model Predictive Control (MPC) has garnered significant attention in recent years due to its ability to predict future states and optimize trajectories under given constraints~\cite{GARCIA1989}. These algorithms can provide optimal solutions under certain conditions but may be computationally intensive and less effective in highly dynamic or unpredictable environments.
Graph-based methods~\cite{Zhou2022,Teng2023} discretize the planning space into nodes and edges, symbolizing possible vehicle positions and paths. This approach is valid in structured and well-known environments like road networks. This enables efficient path planning through well-established graph algorithms. Unfortunately, these algorithms can be limited by the accuracy of the graph representation and may not handle dynamic obstacles well since the spatial path does not consider speed or time. A subsequent velocity planning step incorporates temporal information by assigning speeds and timing to the path, resulting in a spatiotemporal trajectory. This methodology uses advanced planning algorithms, such as time-dependent A*, spatiotemporal graphs, and time-dependent reachability graphs, to identify the shortest or most feasible path~\cite{Rowold2022,Stahl2019,Huang2017,Gasparetto2015,Xin2021}.
Sampling-based methods ~\cite{werling.2012,werling.2010, Liang2015, Huang2023} have gained popularity for their effectiveness in high-dimensional spaces. These methods randomly search the free space around the vehicle's current state for target states. Once an available target state is found, a trajectory is created which connects the current driving state of the vehicle to the selected target state. These algorithms are generally efficient by introducing randomness but do not provide guarantees for finding an optimal trajectory.
Machine learning-based methods~\cite{Chen2019,Schoemer2020} introduce learning techniques, often utilizing neural networks to adapt and improve planning strategies based on past experiences (data or predictions) or simulated scenarios. These algorithms can adapt to new situations, learn from experience, and behave more effectively in dynamic and unpredictable environments. The most important categories are imitation learning and reinforcement learning. Imitation learning involves training a model by mimicking expert behavior, whereas reinforcement learning focuses on learning optimal actions through trial-and-error interactions with the environment to maximize cumulative rewards~\cite{Aradi2022,Elallid2022,Dong2023,Muhammad2021,bojarski2016end}. Unfortunately, imitation learning demands a substantial amount of data, while reinforcement learning requires a detailed description of the environment to be transferable to real-world applications. Additionally, these algorithms often lack interpretability~\cite{Aradi2022}. For this reason, hybrid models are often used to improve the training process and the algorithm's applicability~\cite{Klimke2023,Albarella2023,Ichter2018}.


\section{Methodology}
\label{sec:method}
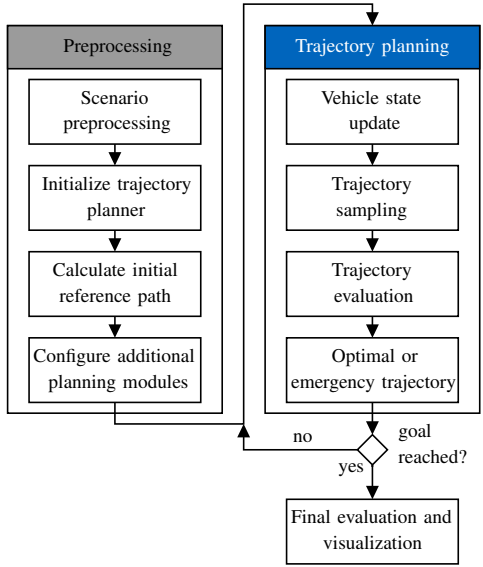
\begin{figure}[ht!]
    \centering
    \resizebox{0.38\textwidth}{!}{
    \tikzsetnextfilename{PlanningProcedure}
\tikzset{every picture/.style={line width=0.75pt}} 

\begin{tikzpicture}[x=1pt,y=1pt,yscale=-1,xscale=1]

\draw    (280,105) -- (280,114.02) ;
\draw [shift={(280,115)}, rotate = 270] [fill={rgb, 255:red, 0; green, 0; blue, 0 }  ][line width=0.08]  [draw opacity=0] (6.25,-3) -- (0,0) -- (6.25,3) -- cycle    ;
\draw    (280,145) -- (280,152) ;
\draw [shift={(280,155)}, rotate = 270] [fill={rgb, 255:red, 0; green, 0; blue, 0 }  ][line width=0.08]  [draw opacity=0] (6.25,-3) -- (0,0) -- (6.25,3) -- cycle    ;
\draw   (350,50) -- (450,50) -- (450,230) -- (350,230) -- cycle ;
\draw    (280,225) -- (280,235) -- (340,235) -- (340,40) -- (400,40) -- (400,47) ;
\draw [shift={(400,50)}, rotate = 270] [fill={rgb, 255:red, 0; green, 0; blue, 0 }  ][line width=0.08]  [draw opacity=0] (6.25,-3) -- (0,0) -- (6.25,3) -- cycle    ;
\draw  [color={rgb, 255:red, 0; green, 0; blue, 0 }  ,draw opacity=1 ] (240,75) -- (320,75) -- (320,105) -- (240,105) -- cycle ;
\draw  [color={rgb, 255:red, 0; green, 0; blue, 0 }  ,draw opacity=1 ] (240,115) -- (320,115) -- (320,145) -- (240,145) -- cycle ;
\draw    (280,185) -- (280,192) ;
\draw [shift={(280,195)}, rotate = 270] [fill={rgb, 255:red, 0; green, 0; blue, 0 }  ][line width=0.08]  [draw opacity=0] (6.25,-3) -- (0,0) -- (6.25,3) -- cycle    ;
\draw  [color={rgb, 255:red, 0; green, 0; blue, 0 }  ,draw opacity=1 ] (240,155) -- (320,155) -- (320,185) -- (240,185) -- cycle ;
\draw  [color={rgb, 255:red, 0; green, 0; blue, 0 }  ,draw opacity=1 ] (240,195) -- (320,195) -- (320,225) -- (240,225) -- cycle ;
\draw  [color={rgb, 255:red, 0; green, 0; blue, 0 }  ,draw opacity=1 ] (360,75) -- (440,75) -- (440,105) -- (360,105) -- cycle ;
\draw  [color={rgb, 255:red, 0; green, 0; blue, 0 }  ,draw opacity=1 ] (360,115) -- (440,115) -- (440,145) -- (360,145) -- cycle ;
\draw    (400,105) -- (400,112) ;
\draw [shift={(400,115)}, rotate = 270] [fill={rgb, 255:red, 0; green, 0; blue, 0 }  ][line width=0.08]  [draw opacity=0] (6.25,-3) -- (0,0) -- (6.25,3) -- cycle    ;
\draw  [color={rgb, 255:red, 0; green, 0; blue, 0 }  ,draw opacity=1 ] (360,155) -- (440,155) -- (440,185) -- (360,185) -- cycle ;
\draw    (400,145) -- (400,152) ;
\draw [shift={(400,155)}, rotate = 270] [fill={rgb, 255:red, 0; green, 0; blue, 0 }  ][line width=0.08]  [draw opacity=0] (6.25,-3) -- (0,0) -- (6.25,3) -- cycle    ;
\draw  [color={rgb, 255:red, 0; green, 0; blue, 0 }  ,draw opacity=1 ] (360,195) -- (440,195) -- (440,225) -- (360,225) -- cycle ;
\draw    (400,185) -- (400,190) -- (400,192) ;
\draw [shift={(400,195)}, rotate = 270] [fill={rgb, 255:red, 0; green, 0; blue, 0 }  ][line width=0.08]  [draw opacity=0] (6.25,-3) -- (0,0) -- (6.25,3) -- cycle    ;
\draw    (400,225) -- (400,237) ;
\draw [shift={(400,240)}, rotate = 270] [fill={rgb, 255:red, 0; green, 0; blue, 0 }  ][line width=0.08]  [draw opacity=0] (6.25,-3) -- (0,0) -- (6.25,3) -- cycle    ;
\draw   (400,240) -- (407,247) -- (400,254) -- (393,247) -- cycle ;
\draw    (393,247) -- (340,247) -- (340,238) ;
\draw [shift={(340,235)}, rotate = 90] [fill={rgb, 255:red, 0; green, 0; blue, 0 }  ][line width=0.08]  [draw opacity=0] (6.25,-3) -- (0,0) -- (6.25,3) -- cycle    ;
\draw    (400,255) -- (400,254) -- (400,267) ;
\draw [shift={(400,270)}, rotate = 270] [fill={rgb, 255:red, 0; green, 0; blue, 0 }  ][line width=0.08]  [draw opacity=0] (6.25,-3) -- (0,0) -- (6.25,3) -- cycle    ;
\draw  [color={rgb, 255:red, 0; green, 0; blue, 0 }  ,draw opacity=1 ] (360,270) -- (440,270) -- (440,300) -- (360,300) -- cycle ;
\draw   (230,50) -- (330,50) -- (330,230) -- (230,230) -- cycle ;
\draw  [color={rgb, 255:red, 0; green, 0; blue, 0 }  ,draw opacity=1 ][fill={rgb, 255:red, 155; green, 155; blue, 155 }  ,fill opacity=1 ] (230,50) -- (330,50) -- (330,70) -- (230,70) -- cycle ;
\draw  [color={rgb, 255:red, 0; green, 0; blue, 0 }  ,draw opacity=1 ][fill={rgb, 255:red, 0; green, 101; blue, 189 }  ,fill opacity=1 ] (350,50) -- (450,50) -- (450,70) -- (350,70) -- cycle ;

\draw (263,79) node [anchor=north west][inner sep=0.75pt]  [font=\small,color={rgb, 255:red, 0; green, 0; blue, 0 }  ,opacity=1 ] [align=left] {  Scenario };
\draw (255,91) node [anchor=north west][inner sep=0.75pt]  [font=\small,color={rgb, 255:red, 0; green, 0; blue, 0 }  ,opacity=1 ] [align=left] { preprocessing };
\draw (245,119) node [anchor=north west][inner sep=0.75pt]  [font=\small,color={rgb, 255:red, 0; green, 0; blue, 0 }  ,opacity=1 ] [align=left] {Initialize trajectory };
\draw (266,131) node [anchor=north west][inner sep=0.75pt]  [font=\small,color={rgb, 255:red, 0; green, 0; blue, 0 }  ,opacity=1 ] [align=left] {planner};
\draw (250,159) node [anchor=north west][inner sep=0.75pt]  [font=\small,color={rgb, 255:red, 0; green, 0; blue, 0 }  ,opacity=1 ] [align=left] {Calculate initial};
\draw (253,171) node [anchor=north west][inner sep=0.75pt]  [font=\small,color={rgb, 255:red, 0; green, 0; blue, 0 }  ,opacity=1 ] [align=left] {reference path};
\draw (241,199) node [anchor=north west][inner sep=0.75pt]  [font=\small,color={rgb, 255:red, 0; green, 0; blue, 0 }  ,opacity=1 ] [align=left] {Configure additional };
\draw (248,211) node [anchor=north west][inner sep=0.75pt]  [font=\small,color={rgb, 255:red, 0; green, 0; blue, 0 }  ,opacity=1 ] [align=left] {planning modules};
\draw (375,79) node [anchor=north west][inner sep=0.75pt]  [font=\small,color={rgb, 255:red, 0; green, 0; blue, 0 }  ,opacity=1 ] [align=left] {Vehicle state };
\draw (388,91) node [anchor=north west][inner sep=0.75pt]  [font=\small,color={rgb, 255:red, 0; green, 0; blue, 0 }  ,opacity=1 ] [align=left] {update};
\draw (380,119) node [anchor=north west][inner sep=0.75pt]  [font=\small,color={rgb, 255:red, 0; green, 0; blue, 0 }  ,opacity=1 ] [align=left] {Trajectory};
\draw (382,131) node [anchor=north west][inner sep=0.75pt]  [font=\small,color={rgb, 255:red, 0; green, 0; blue, 0 }  ,opacity=1 ] [align=left] {sampling };
\draw (380,159) node [anchor=north west][inner sep=0.75pt]  [font=\small,color={rgb, 255:red, 0; green, 0; blue, 0 }  ,opacity=1 ] [align=left] {Trajectory};
\draw (381,171) node [anchor=north west][inner sep=0.75pt]  [font=\small,color={rgb, 255:red, 0; green, 0; blue, 0 }  ,opacity=1 ] [align=left] {evaluation};
\draw (380,199) node [anchor=north west][inner sep=0.75pt]  [font=\small,color={rgb, 255:red, 0; green, 0; blue, 0 }  ,opacity=1 ] [align=left] {Optimal or};
\draw (361,211) node [anchor=north west][inner sep=0.75pt]  [font=\small,color={rgb, 255:red, 0; green, 0; blue, 0 }  ,opacity=1 ] [align=left] {emergency trajectory};
\draw (362,238) node [anchor=north west][inner sep=0.75pt]  [font=\small,color={rgb, 255:red, 0; green, 0; blue, 0 }  ,opacity=1 ] [align=left] {no};
\draw (383,252) node [anchor=north west][inner sep=0.75pt]  [font=\small,color={rgb, 255:red, 0; green, 0; blue, 0 }  ,opacity=1 ] [align=left] {yes};
\draw (361,274) node [anchor=north west][inner sep=0.75pt]  [font=\small,color={rgb, 255:red, 0; green, 0; blue, 0 }  ,opacity=1 ] [align=left] {Final evaluation and};
\draw (376,286) node [anchor=north west][inner sep=0.75pt]  [font=\small,color={rgb, 255:red, 0; green, 0; blue, 0 }  ,opacity=1 ] [align=left] {visualization};
\draw (363,55) node [anchor=north west][inner sep=0.75pt]   [align=left] {\textcolor{white}{\small Trajectory planning}};
\draw (411,234) node [anchor=north west][inner sep=0.75pt]  [color={rgb, 255:red, 0; green, 0; blue, 0 }  ,opacity=1 ] [align=left] {{\small goal }\\{\small reached?}};
\draw (255,55) node [anchor=north west][inner sep=0.75pt]   [align=left] {{\small Preprocessing}};

\end{tikzpicture}
    }
	\caption{Simplified trajectory planning procedure.}
	\label{fig:planningprocedure}
\end{figure}

This section introduces the FRENETIX algorithm, a motion planning algorithm for AVs. FRENETIX enables efficient, safe, and reliable navigation (e.g., overtaking), especially in dynamic and complex environments. 
The modular structure (\cref{fig:modules}) of FRENETIX is a key feature, simplifying the planning process while enhancing adaptability and scalability for different scenarios. Developed using Python for its flexibility and prototyping capabilities, FRENETIX also incorporates C++ components to improve computational efficiency. This combination results in a system that optimizes performance and is practical for deployment.
At its heart is an iterative motion planning cycle, including sampling, cost functions, risk assessments, validity checks, and safety features for continuously refining the vehicle's trajectory in response to environmental changes. A simplified overview of the trajectory planning procedure is shown in ~\cref{fig:planningprocedure}. The module-based overview of the motion planning framework FRENETIX can be seen in~\cref{fig:modules}. 

\begin{figure}%
    \centering
    \hspace{1mm}
    \begin{tikzpicture}[font=\scriptsize]
        \node[inner sep=0pt] at (0.7,0) {\includegraphics[height=2.5mm]{figures/ego.png}};
        \node[align=left, anchor=west] at (1.0,0) {ego \\ vehicle};
    
        \node[inner sep=0pt] at (2.6,0) {\includegraphics[height=2.5mm]{figures/dyn_obstacle.png}};
        \node[align=left, anchor=west] at (2.9,0) {dynamic \\ obstacle};
    
        \draw[thick, GreenCR] (4.2,0) -- (4.6,0);
        \node[align=left, anchor=west] at (4.6,0) {reference \\ path $\Gamma$};

        \node[inner sep=0pt] at (6.3,0) {\arrowCR};
        \node[align=left, anchor=west] at (6.6,0) {start};

        \node[inner sep=0pt] at (7.7,0) {\boxCR};
        \node[align=left, anchor=west] at (7.9,0) {Goal \\ Area};
        
    \end{tikzpicture}
    \subfloat[][\scriptsize{Initial scenario at timestep 0, depicting the vehicle's starting position and potential trajectories within the lanelet network.}]{\label{fig:scenario_visualization_0} \fbox{\includegraphics[width=0.45\textwidth, trim={7cm 17.5cm 7cm 16.45cm},clip]{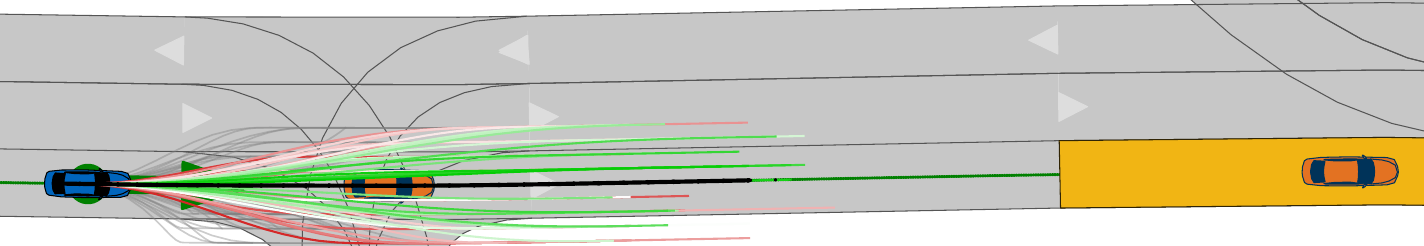}}} \\
    \subfloat[][Progression of the scenario at timestep 35, illustrating dynamic trajectory adjustments based on real-time environmental feedback.]{\label{fig:scenario_visualization_35} \fbox{\includegraphics[width=0.45\textwidth, trim={7cm 17.5cm 7cm 16.45cm},clip]{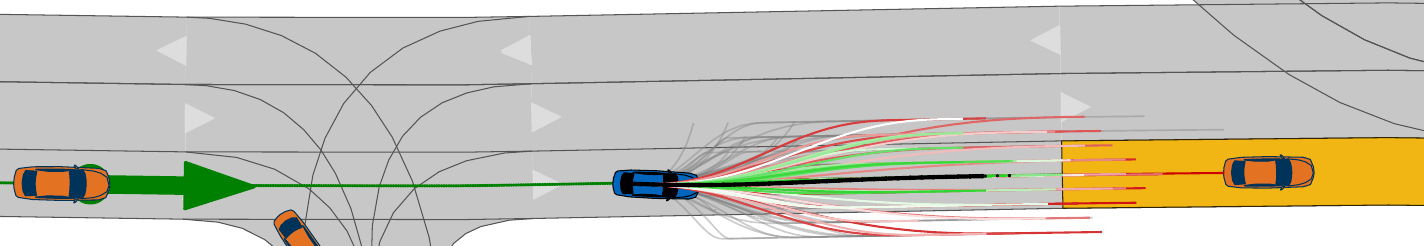}}} \\
    \subfloat[][Final trajectory, demonstrating the optimized path and states the vehicle will follow within the lanelet network.]{\label{fig:scenario_visualization_final} \fbox{\includegraphics[width=0.45\textwidth, trim={7cm 17.5cm 7cm 16.45cm},clip]{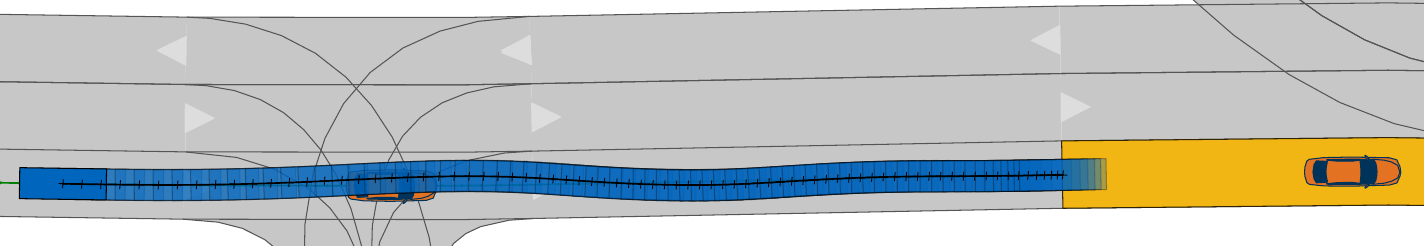}}}%
    \caption{Visualization of the sequential progression of the motion planning algorithm over time. The black trajectory shows the selected optimal trajectory. The gray trajectories are kinematically infeasible. The feasible trajectories are highlighted using a color scale corresponding to the cost, where green indicates a low cost and red indicates a high cost.}%
    \label{fig:cont}%
\end{figure}
 
\subsection{Preprocessing}
\label{subsec:preprocessing}
The trajectory planning algorithm obtains the environment data from the map configuration, in our case, from the CommonRoad~\cite{commonroad,Wuersching2024} scenario database (\cref{fig:cont}). The environment, modeled as a semantic lanelet network with static and dynamic obstacles and traffic signs, contains traffic participant data, regulatory elements, and the specific planning problem. Each obstacle is characterized by its type, dimensions, location, and orientation. The planning problem is defined by the initial state of the ego vehicle and the conditions required to reach a specific goal.
An initial optimal global route through the lanelet network is needed to form a reference path $\Gamma$ for subsequent trajectory generation in the Frenet coordinate system to localize the ego vehicle. To select the optimal sequence of lanelets, a graph optimization algorithm, such as Dijkstra \cite{Dijkstra.1959} or A* \cite{Hart.1968}, is employed. The selected lanelets are interconnected through their centroids, forming a navigable route.

\subsection{Trajectory Planning Cycle}
\label{subsec:trajectoryplanningcycle}
We differentiate between the trajectory planning cycle and the motion planning cycle. While trajectory planning (\cref{fig:planningprocedure}) is limited to the core functionality of trajectory generation and evaluation, motion planning represents the entire FRENETIX module (\cref{fig:modules}) with all modules and extensions. 

\textbf{Vehicle state update:} The trajectory planning cycle starts by updating the ego vehicle's state in our trajectory planning algorithm. This involves refreshing the state vector with the latest position and trajectory data and merging past states with current planner-directed controls.

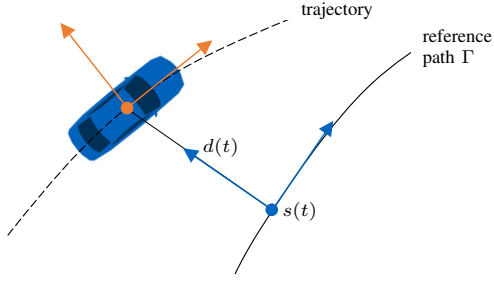
\begin{figure}[!ht]%
    \centering
    \begin{tikzpicture}[font=\scriptsize]
        \node[inner sep=0pt] at (0,0) {\includegraphics[width=0.3\textwidth, trim={0cm 1.5cm 2.5cm 1.5cm},clip]{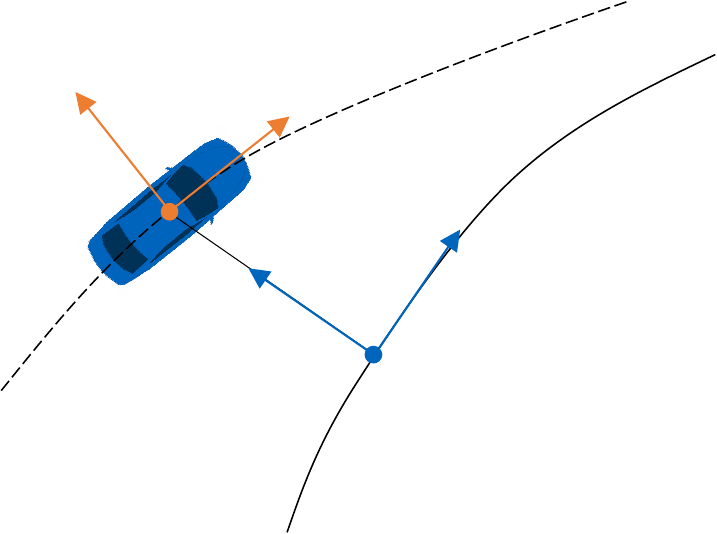}};
        \node[align=left, anchor=west] at (1.1,1.8) {trajectory};
        \node[align=left, anchor=west] at (2.7,1.3) {reference \\ path $\Gamma$};
        \node[align=left, anchor=west] at (0.85,-0.85) {$s(t)$};
        \node[align=left, anchor=west] at (-0.2,0.0) {$d(t)$};
    \end{tikzpicture}
    \caption{Frenet coordinate system for trajectory generation, adapted from \cite{werling.2012}.}
    \label{fig:frenet}
\end{figure}

\textbf{Trajectory sampling:} Our planner is implemented as a semi-reactive approach~\cite{werling.2010} to generate vehicle trajectories by continuously optimizing local paths and performing cyclic replanning steps  (\cref{fig:cont}b). Samples are generated in the Frenet coordinate system to simplify trajectory planning by separating longitudinal and lateral vehicle motion. The vehicle's position in the Frenet frame is quantified using two key parameters: the lateral displacement, denoted as $d$, and the longitudinal displacement, represented by $s$, as illustrated in \cref{fig:frenet}~\cite{werling.2010}. Unlike the traditional Cartesian coordinate system, the Frenet coordinate system adopts a path-focused methodology. Within this system, a vehicle's position and movement are characterized by their relationship to a predetermined reference path, providing a distinct, path-oriented viewpoint. 
The trajectory planning algorithm generates potential final states for lateral and longitudinal trajectories within a predetermined discretization scheme shown in~\cref{tab:sampling_matrix}.
\begin{table}[ht!]
\centering
\caption{Sampling matrix.}
\renewcommand{\arraystretch}{1.3}
\begin{tabular}{l l}
\hline
    Sampling category    & Sampling scheme \\ \hline\hline
    time-sampling        & $[t_{\text{min}},t_1,t_2,...,t_{\text{max}}]$   \\ 
    d-sampling           & $[d_{\text{min}},d_1,...,d_{\text{curr}},d_{\text{curr}+1},...,d_{\text{max}}]$   \\ 
    velocity-sampling    & $[v_{\text{min}},v_1,...,v_{\text{curr}},v_{\text{curr}+1},...,v_{\text{max}}]$   \\ 
    s-sampling           & $[s_{\text{curr}},s_1,s_2,...,s_{\text{max}}]$   \\
\hline
\end{tabular}
\label{tab:sampling_matrix}
\end{table}

The density and number of trajectories can be defined. $t_{\text{max}}$ corresponds to the maximum sampled time horizon $\tau$ and the index $curr$ represents a clipping to the current ego vehicle state. All trajectories are extended to a fixed planning horizon $T$ to ensure comparability between all samples for later cost evaluations. 
$v_{\text{min}}$ and $v_{\text{max}}$ are dependent on the current vehicle state and the acceleration limits. Due to the dependencies, the s-sampling method is only used if velocity-sampling is deactivated. This is because the longitudinal end state can only be defined by the velocity or position range to avoid overdetermination. The sampling scheme in \cref{tab:sampling_matrix} contains the possible variants and the number of trajectories to be generated. There are extensions to the systematic sampling procedure, e.g., limiting sampling to the width of the roadway or the reachable set of the ego vehicle~\cite{Manzinger2021}.
For lateral trajectories, the sampling process involves considering different lateral distances $d_\mathrm{\tau}$ from the reference path $\Gamma$ and distinct time horizons $\tau$. Similarly, for longitudinal trajectories, the algorithm samples endpoints at various velocities $v_\mathrm{\tau}$ and again distinct time horizons. Alternatively, the endpoint $s_\mathrm{\tau}$ can be defined directly over all sampled time horizons $\tau$. In this case, the velocity profile is adapted to the endpoint constraint. This method ensures that the longitudinal trajectory samples represent different speeds and acceleration profiles.
These sampled final states serve as targets for the vehicle to conclude its trajectory. To effectively connect an initial state $\zeta(0) = \zeta_\mathrm{0}$ with a final state $\zeta(\tau) = \zeta_\mathrm{\tau}$ and form a coherent path, suitable polynomial functions are employed (\cref{fig:cont}c).
In lateral motion planning, quintic polynomial functions are utilized to guarantee a trajectory that is both smooth and minimal in jerk \cite{Takahashi}. Consequently, the trajectory along the lateral direction is represented by a polynomial function of the form:
\begin{equation}
    \zeta(t) = c_\mathrm{0} + c_\mathrm{1} \, t + c_\mathrm{2} \, t^{2} + c_\mathrm{3} \, t^{3} + c_\mathrm{4} \, t^{4} + c_\mathrm{5} \, t^{5}
    \label{eq:quinticpolynom}
\end{equation}

Differentiating the polynomial function (\cref{eq:quinticpolynom}) twice yields a system of equations as follows:
\begin{equation}
    \begin{bmatrix}
        \zeta(t) \\
        \dot{\zeta}(t) \\
        \ddot{\zeta}(t) \\
    \end{bmatrix} = 
    \begin{bmatrix}
        1 & t & t^2 & t^3 & t^4 & t^5 \\
        0 & 1 & 2t & 3t^2 & 4t^3 & 5t^4 \\
        0 & 0 & 2 & 6t & 12t^2 & 20t^3 \\
    \end{bmatrix}
    \begin{bmatrix}
        c_0 \\
        c_1 \\
        c_2 \\
        c_3 \\
        c_4 \\
        c_5 \\
    \end{bmatrix}
    \label{polynom}
\end{equation}

This matrix equation establishes the relationship between the polynomial coefficients \( c_0, \ldots, c_5 \) and the vehicle's state. In the case of the lateral movement, the boundary conditions for solving \( c_0, \ldots, c_5 \) are set as follows \cite{werling.2012}:
\begin{equation}
    \begin{aligned}
    \zeta(t = 0) &= d_0, & \zeta(t = \tau) &= d_\mathrm{\tau}, \\
    \dot{\zeta}(t = 0) &= \dot{d}_0, & \dot{\zeta}(t = \tau) &= \dot{d}_\mathrm{\tau} = 0, \\
    \ddot{\zeta}(t = 0) &= \ddot{d}_0, & \ddot{\zeta}(t = \tau) &= \ddot{d}_\mathrm{\tau} = 0.
    \end{aligned}
    \label{eq:quinticmatrix}
\end{equation}

\begin{figure*}[ht!]
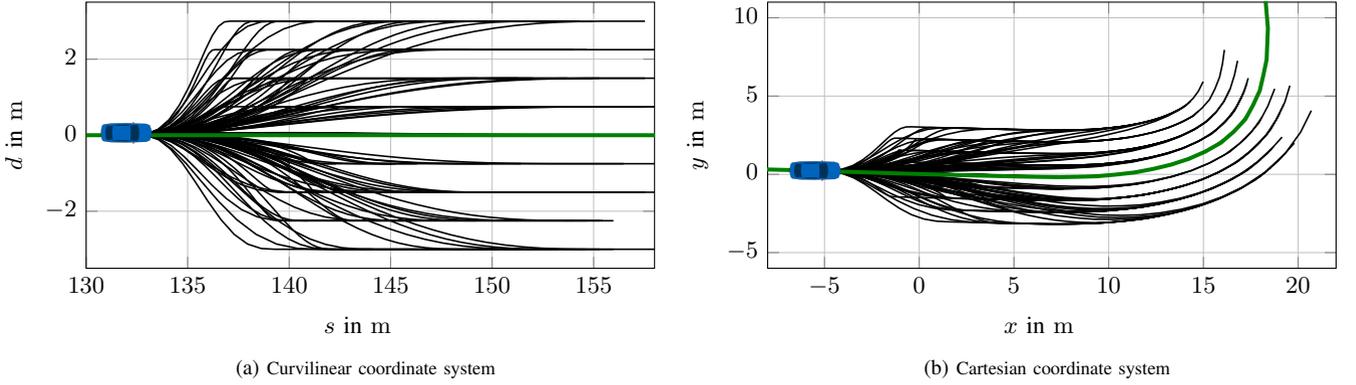

    \centering
    \subfloat[][\scriptsize{Curvilinear coordinate system}]{\label{fig:sampling_curvilinear} \input{figures/sampling/TrajectorySampling_curv}} \hspace{1.0cm}
    \subfloat[][\scriptsize{Cartesian coordinate system}]{\label{fig:sampling_cartesian} \input{figures/sampling/TrajectorySampling_cart}} \\
    \caption{Sampling process in different coordinate systems, as per \cite{werling.2010}: subfigure (a) illustrates trajectories in curvilinear coordinates, capturing the primary direction of movement within a curved reference frame. Subfigure (b) translates these trajectories into cartesian coordinates. The ego vehicle is blue (\protect\egoCRsmall), indicating its position following the sampled trajectories along the reference path (\protect\refpathCR). }
    \label{fig:sampling}
\end{figure*}

The final lateral state velocity $\dot{d}_\mathrm{\tau}$ and acceleration $\ddot{d}_\mathrm{\tau}$ are set to zero, as the planning algorithm aims for a movement parallel to the reference path. 
For trajectory generation in the longitudinal direction, quartic polynomials are employed, providing an adequate description of vehicle motions in the longitudinal direction while ensuring minimal jerk \cite{werling.2012}. 
When s-sampling is enabled, quintic polynomials are used because the endpoint manifold is omitted.
The quartic polynomials in the longitudinal direction enable a manifold of the end positions since the velocity at the endpoint is not zero, while the acceleration is. The coefficients can analogously be calculated by solving an adapted version of \cref{eq:quinticmatrix} and applying the appropriate boundary conditions presented in \cite{werling.2012}. 
By sampling $m$ trajectories in the longitudinal direction and $n$ in the lateral direction, a set comprising $m \times n$ trajectories is constructed through a systematic crosswise superposition of these two sets. 
The number of trajectories generated depends on the sampling scheme and density. With the completion of the sampling process, the trajectory samples are transformed back into global cartesian coordinates for further evaluation steps. The sampling process is visualized in ~\cref{fig:sampling}, where the resulting samples are shown in curvilinear and cartesian coordinate systems.

\textbf{Trajectory evaluation:} The trajectory evaluation stage is required to assess the trajectory samples w.r.t. feasibility and optimality. This stage adopts a systematic funnel approach, processing all trajectory samples generated in the preceding step through a sequence of evaluations. These assessments are designed to identify the optimal trajectory that the vehicle should adhere to for the subsequent timestep.

\textit{1.) Kinematic check:} The first step in the evaluation process is a kinematic check. This involves assessing each trajectory sample to ensure it satisfies the kinematic constraints of the vehicle based on a kinematic single-track model. This step ensures that the trajectories are within the physical movement capabilities of the ego vehicle, considering the acceleration, curvature, curvature rate, and yaw rate.
The permissible acceleration $ a_{\text{permissible}}(t) $ is expressed as~\cite{commonroad}:
\begin{equation}
a_{\text{permissible}}(t) = 
  \begin{cases} 
   a_{\text{max}} \cdot \frac{v_{\text{switch}}}{v(t)} & \text{if } v(t) > v_{\text{switch}} \\
   a_{\text{max}} & \text{if } v(t) \leq v_{\text{switch}}
  \end{cases}
\end{equation}

Let $ v(t) $ denote the vehicle's velocity at any point along the trajectory, and $ a(t) $ represent the corresponding acceleration. Given a predefined maximum acceleration $ a_{\text{max}} $ and a threshold velocity $ v_{\text{switch}} $, which delineates the transition from constant to variable acceleration limits. The kinematic feasibility is thus determined by evaluating whether the actual acceleration $ a(t) $ for each trajectory point remains within the bounds of $ -a_{\text{max}} $ and $ a_{\text{permissible}}(t) $:
\begin{equation}
    -a_{\text{max}} \leq a(t) \leq a_{\text{permissible}}(t), \quad \forall t \in [t_0, t_f]
\end{equation}

Here, $ t_0 $ and $ t_f $ represent the start and end of the considered time interval, respectively. 
Furthermore, the curvature $\kappa(t)$ of the trajectory at any point in time must not exceed the maximum curvature $\kappa_{\text{max}}$, which is derived from the maximum allowable steering angle $\delta_{\text{max}}$ and the wheelbase $L$ of the vehicle:
\begin{equation}
    \label{eq:curvature}
    \kappa_{max} = \frac{\tan(\delta_{max})}{L} 
\end{equation}
Thus, the curvature constraint can be stated according to \cref{eq:curvatureconstraint}.
\begin{equation}
    \label{eq:curvatureconstraint}
    |\kappa(t)| \leq \kappa_{max}, \quad \forall t \in [t_0, t_f] 
\end{equation}

\begin{table*}[ht]
    \newlength{\myvspace}
    \setlength{\myvspace}{1.2mm}
	\centering
    \caption{Implemented Cost Functions.}
	\label{tab:costfunctions}
	\begin{tabularx}{0.92\textwidth}{p {0.2cm} p{2.0cm} p{4.8cm} p{8cm}}
		\toprule
		& \textbf{Cost function} & \textbf{Formula} & \textbf{Explanation}\\
		\midrule
		& \multirow{2}{*}{Acceleration} & \multirow{2}{*}{$J_\mathrm{A} = \int_{t_\mathrm{0}}^{t_\mathrm{f}} a^{2} \mathrm{d}t$} & quantifies the total squared acceleration $a$, penalizing large accelerations \vspace{\myvspace} \\
		& Jerk & $J_\mathrm{J} = \int_{t_\mathrm{0}}^{t_\mathrm{f}} \dot{a}^{2} \mathrm{d}t$ & quantifies the total squared jerk $\dot{a}$, penalizing abrupt changes \vspace{\myvspace} \\
        & \multirow{2}{*}{Lateral jerk} & \multirow{2}{*}{$J_\mathrm{J_\mathrm{lat}} = \int_{t_\mathrm{0}}^{t_\mathrm{f}} \dot{a}_\mathrm{lat}^{2} \mathrm{d}t$} & quantifies the total squared lateral jerk $\dot{a}_\mathrm{lat}$, penalizing sudden changes in lateral acceleration \vspace{\myvspace} \\
        \multirow{-7}{*}{\settowidth\rotheadsize{Comfort}\rotcell{Comfort}} & \multirow{2}{*}{Longitudinal jerk} & \multirow{2}{*}{$J_\mathrm{J_\mathrm{lon}} = \int_{t_\mathrm{0}}^{t_\mathrm{f}} \dot{a}_\mathrm{lon}^{2} \mathrm{d}t$} & quantifies the total squared longitudinal jerk $\dot{a}_\mathrm{lat}$, penalizing sudden changes in longitudinal acceleration \vspace{\myvspace} \\
        \midrule \vspace{\myvspace}
         \multirow{-3}{*}{\settowidth\rotheadsize{.Efficiency..}\rotcell{\phantom{.} Efficiency}} & \multirow{3}{*}{Velocity offset} & \multirow{3}{4.8cm}{$J_\mathrm{VO} = \int_{t_\mathrm{s}}^{t_\mathrm{f}} \abs{v(t) - v_\mathrm{ref}(t)} \, \mathrm{d}t \newline \phantom{bbbbbbb} + (v(t_\mathrm{f}) - v_\mathrm{ref}(t))^{2}$} & calculates the absolute velocity offset compared to a reference velocity $v_\mathrm{ref}(t)$ over a given period from $t_\mathrm{s}$ to $t_\mathrm{f}$, with an additional emphasis on the squared difference in velocity at the final time $t_\mathrm{f}$ \vspace{\myvspace} \\
        \midrule \vspace{\myvspace}
        & \multirow{2}{*}{Dist. to ref. path} & \multirow{2}{*}{$J_\mathrm{RP} = \int_{t_\mathrm{0}}^{t_\mathrm{f}} {d}^{2}(t) \, \mathrm{d}t$} & measures the total squared distance from the reference path, penalizing deviations from the desired path \vspace{\myvspace} \\
        & \multirow{2}{*}{Dist. to obstacle} & \multirow{2}{*}{$J_\mathrm{DO} = \sum_{1}^{n} \int_{t_\mathrm{0}}^{t_\mathrm{f}} \frac{1}{\Delta x_\mathrm{DO}^2} \mathrm{d}t$} & computes the sum of the inverse squared distances to obstacles $\Delta x_\mathrm{DO}$ for $n$ different obstacles \vspace{\myvspace} \\
        & \multirow{3}{2.0cm}{Collision prob.} & \multirow{3}{*}{$J_\mathrm{CP} = \sum_{1}^{n} \int_{t_\mathrm{0}}^{t_\mathrm{f}} \int f(x, 0, \Sigma_\mathrm{rot}) \mathrm{d}x \, \mathrm{d}t$} & calculates the total collision probability over time for $n$ obstacles by integrating the probability density function with a rotated covariance matrix $\Sigma_\mathrm{rot}$ across a spatial domain $x$ \vspace{\myvspace} \\
        \multirow{-8}{*}{\settowidth\rotheadsize{Safety}\rotcell{Safety}} & \multirow{4}{2.0cm}{Collision prob. mahalanobis} & \multirow{4}{*}{$J_{\mathrm{CM}} = \sum_{1}^{n} \int_{t_{0}}^{t_{f}} \frac{1 - \frac{t}{T}}{d_{\mathrm{M}}(u(t), v(t), \Sigma(t))} \, \mathrm{d}t $} & calculates the collision probability for $n$ obstacles by integrating the inverse Mahalanobis distance $d_\mathrm{M}$ between the trajectory point $u(t)$ and each obstacle's predicted position $v(t)$, with a linearly decreasing weight over $T$ \\
		\bottomrule
	\end{tabularx}
\end{table*}%

The rate of change of the curvature $\dot{\kappa}(t)$ should also be bounded to ensure smooth transitions in steering. Therefore, we assume a maximum acceptable curvature rate $\dot{\kappa}_{max}$ to determine the trajectory's feasibility, where 
\begin{equation}
    |\dot{\kappa}(t)| \leq \dot{\kappa}_{max}, \quad \forall t \in [t_0, t_f]
\end{equation}
Finally, the yaw rate $\dot{\psi}(t)$, which represents the rate of change of the vehicle's orientation $\psi$, should not exceed the maximum yaw rate $\dot{\psi}_{max}$ determined by $\kappa_{max}$ and the vehicle's velocity $v(t)$:
\begin{equation}
    \dot{\psi}_{max}(t) = \kappa_{max} \cdot v(t)
\end{equation}
The yaw rate constraint is:
\begin{equation}
|\dot{\psi}(t)| \leq \dot{\psi}_{max}(t), \quad \forall t \in [t_0, t_f]
\end{equation}
Should any trajectory sample violate these constraints, it is deemed infeasible, signifying a deviation from the vehicle's kinematic capabilities. 

\textit{2.) Cost calculation:} After the kinematic validation, each trajectory is assigned a cost based on a cost function. The cost computation for the trajectory planning algorithm is formulated as a weighted sum of various partial cost components, each quantifying distinct aspects of the trajectory's quality. The total cost $J_\mathrm{sum}(\xi|f_{\xi})$ for a given trajectory $\xi \in \mathcal{T}$ is expressed as:
\begin{equation}
    J_\mathrm{sum}(\xi|f_{\xi}) = \sum_{i=1}^{n} \omega_\mathrm{i} \cdot J_\mathrm{i}(\xi)
\end{equation}
where $J_\mathrm{i}(\xi)$ represents the $i^{th}$ cost function from the implemented set of cost functions. The weighting factor $\omega_\mathrm{i}$ indicates the relative importance of the corresponding cost component in the overall cost calculation. The cost functions encompass criteria such as comfort, efficiency, and particularly safety \cite{Naumann.2020}. Safety is quantified primarily through collision probability costs, derived from predictive models of other traffic participants' movements \cite{walenet}. By estimating the likelihood of a collision based on these predicted trajectories, the algorithm assigns a cost that reflects the potential collision probability. 
Each cost function is designed to provide a robust indicator of the trajectory's viability. An overview of all implemented cost functions is given in \cref{tab:costfunctions}. After the cost computation for each trajectory, they are sorted in ascending order of their respective costs, placing the most cost-effective trajectory at the forefront for further evaluation.

\textit{3.) Collision check:} The sorted trajectories then undergo a collision check using the drivability checker~\cite{DrivabilityChecker}. This step analyzes the trajectories for collisions with static and dynamic objects. 
As trajectories are required to be free of collisions for all continuous times $t$, it is crucial to ensure collision avoidance not just at discrete steps but also between any two subsequent states  $x_1 = x(t_1)$ and $x_2 = x(t_2)$, where \( t_1 \leq t_2 \). Therefore, the collision checker uses an oriented bounding box (OBB) around the occupied spaces for two consecutive time steps. This approach makes it possible to ensure a continuous collision-free path within these intervals~\cite{DrivabilityChecker}.
Initially, the first trajectory in the list is checked for collisions. If it is collision-free, the process stops. As the trajectories are already sorted according to their cost, we only need to find the first collision-free trajectory, significantly reducing the computation time for collision checking.

\textit{4.) Road boundary check:} The final step involves a road boundary check, where only the first collision-free trajectory is initially evaluated for adherence to road boundaries. If it keeps the vehicle on the road, it is accepted; otherwise, the next trajectory is checked until one meets both non-collision and road adherence criteria~\cite{DrivabilityChecker}.

\indent \textbf{Optimal trajectory:}
The output of this evaluation funnel is the optimal trajectory, which has successfully passed through all the assessment layers with the lowest associated costs while ensuring safety. This trajectory is deemed the most suitable for execution by the vehicle, balancing efficiency, safety, and comfort.

\indent \textbf{Emergency risk trajectory:}
If the evaluation does not result in an optimal trajectory, FRENETIX calculates an additional emergency trajectory. This is done by assessing the risk $R(\mathcal{\xi})$ of an unavoidable collision trajectory for the ego vehicle and the potential collision partner. The risk is defined by the maximum harm and the collision probability~\cite{geisslingerconcept}.
\begin{equation}
\begin{split}
	R(\mathcal{\xi}) = \mathrm{max}(p(\mathcal{\xi}) \cdot H(\mathcal{\xi}))
\end{split}
\label{eq:trajrisk}
\end{equation} 
\indent \textbf{Emergency stopping trajectory:} In scenarios where calculating the minimum risk trajectory proves infeasible, for instance, when no prediction information is available, the system defaults to a stopping trajectory. To optimize the braking process and minimize braking distance, the vehicle maintains its current lateral distance ($\text{argmin}_i \, |d_{\text{curr}} - d_{\text{samp}_d[i]}|$) from the reference path. This approach is specifically designed to maximize the absorption of longitudinal negative acceleration during braking. Furthermore, only dynamically feasible trajectories $\mathcal{T}_f$ are considered in this process, ensuring both effectiveness and safety in vehicle maneuvering.

\subsection{Modular Architecture}
\label{sec:modular}
\begin{figure*}[ht!]
    \centering
    \includegraphics[width=0.95\textwidth]{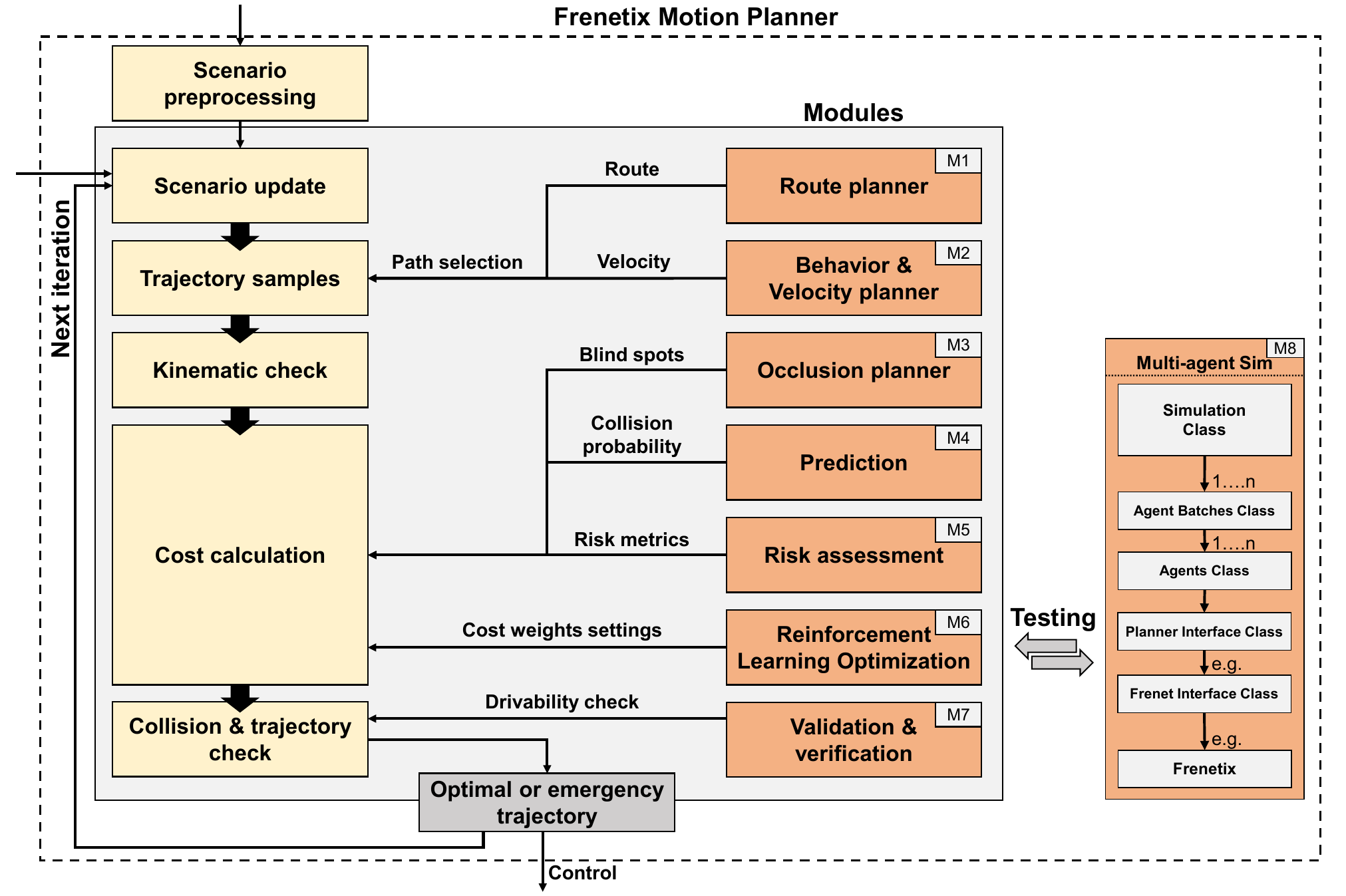}
    \caption{Frenetix Motion Planner planning cycle iteration and module integration.}
    \label{fig:modules}
\end{figure*}

The modular software's main components and core functionalities are shown in~\cref{fig:modules}.
Since FRENETIX has a modular approach, one can initialize further FRENETIX modules to expand the planner's capabilities. The modules influence the motion planning process in each planning iteration. The modules can but do not have to be used. For example, the route may be reviewed and adjusted at each time step, or it may be possible to use only an initial route without further adjustment. 
The exchangeable integrated modules are labeled by \textbf{M1-M8} in \cref{fig:modules}.

\textbf{M1}: The global route planner generates an optimal route to the goal region in Cartesian coordinates\footnote{\url{https://gitlab.lrz.de/tum-cps/commonroad-route-planner}}~\cite{Hart.1968,Dijkstra.1959}. The route determines the sequence of lanelets to be traveled through. The sequence of Cartesian coordinates $[[x_1,y_1],...,[x_n,y_n]]$ is used as an interface from which a smoothed continuous path is calculated. We are calculating a smooth, approximating spline curve for a set of points in multi-dimensional space. By providing the coordinates of these points, the method generates a B-spline representation that best fits the given data~\cite{dierckx1982algorithms,dierckx1981algorithms,dierckx1993curve}. We then discretize the path again to use the trajectory calculation.

\textbf{M2}: The behavior \& velocity planner can generate new potential routes, choose the optimal one, plan the target velocity, and facilitate high-level decision-making. The target speed $V_T$ is calculated depending on the traffic rules, the vehicles in the surrounding area, the driving mode, and the goal features. The selected route, e.g., the route on the highway, can be adjusted to reach the destination according to the set requirements.

\textbf{M3}: Occlusion-aware trajectory planning addresses the risks posed by blind spots~\cite{Trauth2023}. Occluded areas introduce uncertainties that can result in significant personal harm if not accounted for in the planning process. These uncertainties can be effectively integrated into trajectory planning by incorporating phantom objects or risk assessments. This integration can occur either during cost calculation or during the trajectory validation stage, enhancing safety and reliability.

\textbf{M4}: Vehicle trajectory prediction with uncertainty integration~\cite{walenet}; The interface provides a time-based prediction of the position and speed of the surrounding objects. In addition, covariances can be added to calculate the costs more accurately when selecting the trajectory. We utilize the standardized ROS2 message format to establish the interface between the trajectory planning and prediction algorithm~\cite{Ros2}.

\textbf{M5}: Risk assessment and harm estimation of predicted traffic participants~\cite{geisslingerconcept} are crucial in ethical trajectory selection. The analysis distinguishes between vulnerable and non-vulnerable road users. Risk calculations are performed for both the ego vehicle and third-party road users. These risk assessments can inform cost calculations and serve as a trajectory validity check to ensure risk levels do not exceed acceptable thresholds.

\textbf{M6}: Optimizing the trajectory selection process through a reinforcement learning-based framework~\cite{Trauth-RL} addresses the challenge of setting appropriate weightings $[w_1, w_2, \ldots, w_n]$ in sampling-based approaches, particularly in dynamic situations. This hybrid approach enables continuous adaptation to various scenarios by reviewing and adjusting weightings at each time step. Consequently, it enhances the assessment of situations and the recognition of priorities, thereby improving safety. Module \textbf{M8} can be used to train and validate the algorithm within an agent-based simulation framework~\cite{kaufeld2024investigating}.

\textbf{M7}: Validation \& verification of the most cost-efficient trajectories through static and dynamic collision checks\footnote{\url{https://github.com/CommonRoad/commonroad-drivability-checker}}~\cite{DrivabilityChecker}. The module ensures that planned trajectories are both collision-free and road-compliant. It supports geometric shapes and hierarchical representations to enhance collision detection efficiency. With robust interfaces for Python and C++ implementations, it is ideal for real-time applications and simulations. The interface ensures that any CommonRoad trajectory can be processed~\cite{commonroad}.

\textbf{M8}: The Multi-agent simulation framework to investigate and test the motion planning framework before real-world application~\cite{kaufeld2024investigating}. The multiagent simulation module facilitates the use of simulations for various purposes, allowing for the execution of multiple simulations and managing numerous agents concurrently. Within this framework, the simulation class plays a pivotal role in organizing scenario-specific and planning-centric information, including prediction information of other road users. It generates batches of agents to accelerate the computation time, where each batch can comprise an arbitrary number of agents. Parallelization at the highest level is usually the most efficient. The batches enable such parallelization. This agent can be parallelized at the planner level when executing a single agent. These agent instances handle all relevant data and statuses relating to the ego vehicle, i.e., the vehicle directly controlled in the simulation.
A designated planner interface governs the agents' motion planning process. This interface is designed to ensure compatibility with any planning algorithm, thus providing flexibility in assigning different planning algorithms to individual agents as required. At its heart, the framework's primary functions include transferring new individual information to the planning algorithm and the sequential execution of the planner at each time step. Furthermore, it establishes universal interfaces, facilitating access to essential data such as trajectory information and global path details in the CommonRoad format~\cite{commonroad}. Each agent can reach the following states during the simulation according to the CommonRoad planning problems~\cite{commonroad}:
\begin{enumerate}
    \item \textbf{Idle:} The system is inactive or has not yet been assigned a task.
    \item \textbf{Running:} The system actively works towards the goal.
    \item \textbf{Goal Reached:} The goal was successfully achieved within the expected time frame.
    \item \textbf{Goal Reached Outside Target Time:} The goal was achieved later than the specified time frame.
    \item \textbf{Goal Reached Faster Than Target Time:} The goal was achieved faster than the specified time frame.
    \item \textbf{Missed the Target:} The intended goal was not achieved.
    \item \textbf{Time Limit Reached:} The operation has hit its time cap without achieving the goal region.
    \item \textbf{Error:} The system encountered a malfunction or unexpected issue.
    \item \textbf{Collision:} The system experienced a physical collision during operation.
\end{enumerate}
In our specific implementation, we use the Frenet interface to execute the planning phases of the FRENETIX Motion Planner; however, this work focuses on the execution of the single agent. The work of Kaufeld et al.~\cite{kaufeld2024investigating} provides more information on running multi-agent simulations.
In addition, other modules, such as the global planner or the behavior planner, can be integrated into the agent's execution step independently of the used local planning algorithm by processing this information separately by the planning interface.


\section{Results \& Analysis}
\label{sec:results}
In this section, we will first examine the algorithm in simulation before demonstrating its applicability in a real vehicle. The aim is to examine the framework and the trajectory planner as standalone tools rather than comparing them to specific alternatives. This is because their performance can be continually enhanced through individual extensions.

\subsection{Environment \& Evaluation}
\begin{figure}[ht!]
    \centering
    \tikzsetnextfilename{success_rate}
\pgfplotsset{scaled x ticks=false}
\pgfplotsset{scaled y ticks=false}
\begin{tikzpicture}[trim axis left, trim axis right, font=\footnotesize]
    \begin{axis}[
        /pgf/number format/.cd,
        1000 sep={},
        width  = 0.45*\textwidth,
        height = 6cm,
        nodes near coords,
        major x tick style = transparent,
        ybar=2*\pgflinewidth,
        bar width=25pt,
        ymajorgrids = true,
        ylabel = {Number of scenarios},
        symbolic x coords={Success,Collision,No feasible solution,Timelimit},
        xtick = data,
        ytick = {200, 400, 600, ..., 1600},
        scaled y ticks = false,
        enlarge x limits=0.2,
        ymin=0,
        ymax=1600,
        axis x line*=bottom, 
		axis y line*=left,
        tick label style={/pgf/number format/1000 sep={}},
        xticklabel style={text width=1.2cm, align=center},
   		 ]
   		 
        \addplot[style={black,fill=Bluelight,mark=none}] coordinates {(Success, 1539) (Collision,143) (No feasible solution,39) (Timelimit,29)};
            
    \end{axis}
\end{tikzpicture}
	\caption{Scenario evaluation of 1750 CommonRoad scenarios.}
	\label{fig:success_rate}
\end{figure}
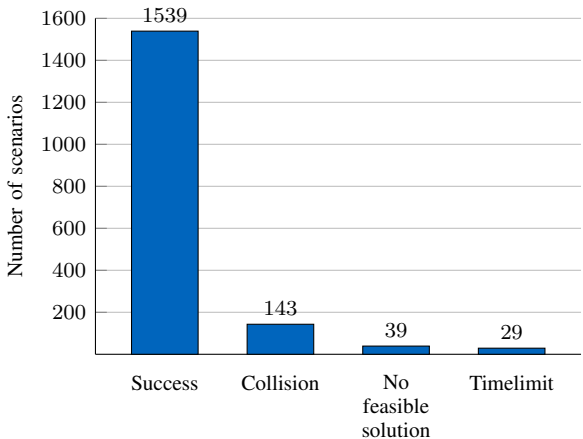

We evaluate our new FRENETIX motion planner in the CommonRoad simulation environment~\cite{commonroad}. The AV has to find a trajectory in given scenarios to reach the goal region in a limited amount of time without a collision and in a kinematically feasible way.
The algorithm's success rate depends on many factors and the difficulty of the scenarios. 
We use \SI{1750}{} CommonRoad scenarios\footnote{\url{https://commonroad.in.tum.de/scenarios}} to evaluate the performance of the algorithm. In this evaluation, we maintain consistent settings and cost weightings for the algorithm, intentionally avoiding adjustments or fine-tuning. The results created are, therefore, from an untuned model lacking specialized behavior features, especially in edge-case scenarios. This ensures that the evaluation of performance is independent and unbiased. \cref{fig:success_rate} shows the simulation results of the \SI{1750}{} scenarios.
Our FRENETIX planner finds in \SI{1539}{} scenarios a safe and valid solution and can solve the scenario. In \SI{143}{} scenarios, FRENETIX is causing a collision. It should be noted that incorrect predictions of vehicle movements and collisions, where the responsibility does not lie with the ego vehicle, cause around \SI{40}{\percent} of the accidents. In \SI{29}{} scenarios, FRENETIX could not reach the goal within a given time limit. However, this only means that the manually designed time limit of the scenario was not met.

\subsection{Vehicle Dynamic Interaction Scenario}
\begin{figure*}[ht!]
    \centering
    \includegraphics[width=1.00\textwidth]{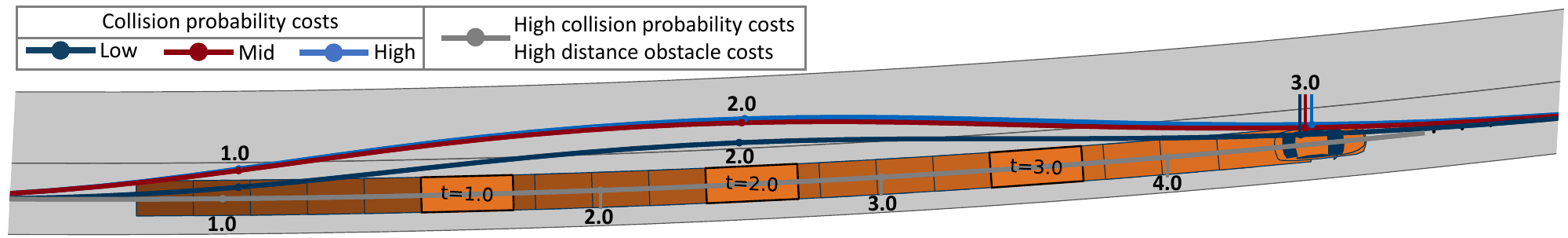}
	\caption{Overtaking scenario with four different cost-weight settings. The numbers next to the trajectories illustrate the progression of the runs in \SI{}{\second}.}
	\label{fig:overtaking}
\end{figure*}

\begin{figure*}[ht!]
    \centering
    \includegraphics[width=1.00\textwidth]{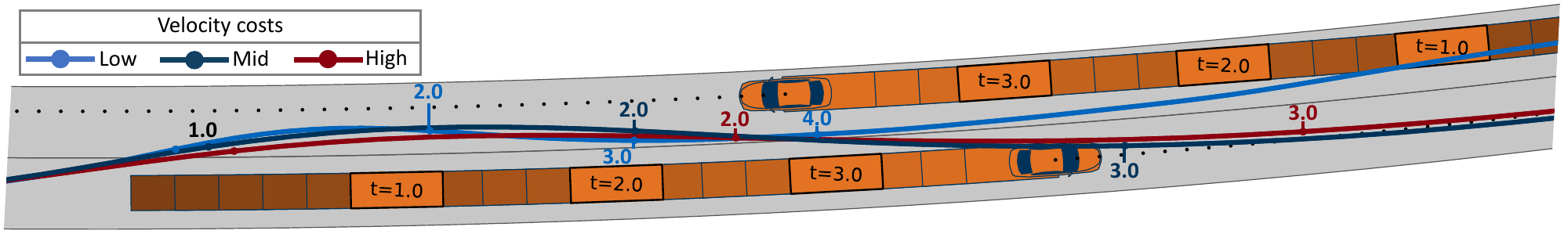}
	\caption{Overtaking scenario with an oncoming vehicle with three different cost-weight settings. The numbers next to the trajectories illustrate the progression of the runs in \SI{}{\second}.}
	\label{fig:overtaking_with_oncoming_traffic}
\end{figure*}

We evaluate FRENETIX in two detailed scenarios: an overtaking maneuver in \cref{fig:overtaking} and an overtaking maneuver with an oncoming vehicle in \cref{fig:overtaking_with_oncoming_traffic}. 
We use different cost weight settings displayed in \cref{tab:cost_weights} to investigate the trajectory selection process while overtaking. 
\begin{table}[ht!]
\centering
\renewcommand{\arraystretch}{1.1}
\caption{Cost weight variations \& planner setting for trajectory selection process.
}
\begin{tabular}{|c|c|}
    \hline
    Planner Settings & Values \\ \hline\hline
    Lateral jerk    & \SI{0.1}{}           \\ 
    Longitudinal jerk    & \SI{0.1}{}   \\
    Distance to reference path    & \SI{0.1}{}   \\  
    Velocity    & [\SI{0.05}{},\SI{0.1}{},\SI{1.0}{}]   \\ 
    Distance to obstacles    & [\SI{0}{},\SI{100}{}]   \\ 
    Collision probability    & [\SI{2}{},\SI{100}{},\SI{1000}{}]   \\ \hline
    Planning horizon $T$ & \SI{3}{\second} \\
    Max./Min. $d$-sampling & [\SI{-3.5}{},\SI{3.5}{}] \\
    \hline
    \end{tabular}
\label{tab:cost_weights}
\end{table}
The weightings depend on the respective cost function, meaning they can only be considered relative to their changes between different runs. In the following investigations, a distinction is made between the different weighting levels to demonstrate the performance and functionality of the algorithm. A detailed parameter analysis is not carried out, as this can be obtained from additional modules~\cite{Trauth-RL}. 
In the first study, without oncoming traffic, the costs of the collision probability varied to investigate the change in driving behavior.
In our study, the overtaken vehicle maintains a constant speed of~\SI{13}{\meter\per\second}. As illustrated in \cref{fig:overtaking}, we observe that augmenting the costs associated with collision probability leads to an increased lateral distance from overtaking vehicle. If the distance is sufficient, the cost of the collision probability decreases so much that it hardly influences the distance so that the vehicle does not move unnecessarily far from the other vehicle. This ensures that the vehicle does not deviate excessively from its reference path. Under these conditions, the overtaking velocity is influenced by the target speed. Conversely, when the costs for increased distance to an obstacle are factored in, the ego vehicle stays behind the preceding vehicle instead of overtaking. This is due to the increased distance to obstacle costs to the leading vehicle, which exceeds the velocity. The trajectory planner can also manage critical situations with conflicting objectives. 
\cref{fig:overtaking_with_oncoming_traffic} shows the same scenario with an oncoming vehicle. The study examines the variation in velocity costs and how to handle the oncoming traffic. The velocity costs are dependent on the target speed. With high-velocity costs, the vehicle accelerates significantly faster and can turn in again sooner after overtaking. With low-velocity costs, the vehicle in front cannot be overtaken in time. The ego vehicle has to pull in again and overtake at a later timestep after \SI{5}{\second}. Nevertheless, none of the different setups results in a collision with the other vehicles. However, the vehicle fails if we deactivate one of the two important cost functions. Collision probability costs, and velocity costs are necessary to achieve the goal region.

\subsection{Calculation time}
To analyze computation times for trajectory planning, the study utilized a Dell Alienware computer equipped with an AMD 7950X processor, 128 GB RAM, and an NVIDIA GeForce RTX 4090 graphics card. We benchmark our FRENETIX C++ implementation against a FRENETIX Python implementation. We distinguish between single-core (SC) and multi-processing (MP) since sampling-based planners are easy to parallelize. Since we are using a sampling scheme, we cannot linearly raise the number of trajectories. We investigate how much time the algorithm needs to create the trajectory samples, check the feasibility of each trajectory, and calculate their costs.
We depict the results of this evaluation in  \cref{tab:calculationtime}.
\begin{table*}[ht!]
    \centering
    \caption{Calculation times (in milliseconds) for trajectory sampling, feasibility check, and cost evaluation.
    }
    \begin{tabular}{|c||c|c|c||c|c|c|}
        \hline
        Trajectories & C++ (SC) & C++ (MP) & \% Difference (SC/MP) & Python (SC) & Python (MP) & \% Difference (SC/MP) \\
        \hline\hline
        50 & \SI{4.68}{\ms} & \SI{3.29}{\ms} & 147.72\% & \SI{35.68}{\ms} & \SI{124.67}{\ms} & 28.62\% \\
        180 & \SI{7.98}{\ms} & \SI{3.92}{\ms} & 203.57\% & \SI{124.69}{\ms} & \SI{133.62}{\ms} & 93.32\% \\
        800 & \SI{29.37}{\ms} & \SI{7.87}{\ms} & 373.73\% & \SI{457.87}{\ms} & \SI{214.72}{\ms} & 213.24\% \\
        3500 & \SI{97.94}{\ms} & \SI{29.82}{\ms} & 328.44\% & \SI{1667.42}{\ms} & \SI{590.36}{\ms} & 282.44\% \\
        13000 & \SI{661.74}{\ms} & \SI{112.38}{\ms} & 588.84\% & \SI{5912.78}{\ms} & \SI{1819.21}{\ms} & 325.02\% \\
        90000 & \SI{5520.05}{\ms} & \SI{717.05}{\ms} & 769.83\% & \SI{44094.33}{\ms} & \SI{10986.14}{\ms} & 401.36\% \\
        \hline
    \end{tabular}  
    \label{tab:calculationtime}
\end{table*}
The results reveal significant differences in calculation times between different implementations. The percentage differences between SC and MP modes were remarkably high, indicating a substantial efficiency gain in multi-processing. For instance, for \SI{90000}{} trajectories, the time was reduced from \SI{5.52}{\second} in SC to \SI{0.717}{\second} in MP. In the comparative analysis of Python implementations, MP exhibits a threshold-dependent efficacy. Specifically, Python-MP demonstrates a net advantage beyond a critical number of trajectories. This is attributable to the inherent overhead associated with MP under Python, which can outweigh the computational gains when dealing with a limited set of trajectories. Conversely, the C++ implementation showcases a more efficient resource utilization. 

\subsection{Real-World Vehicle Deployment}
\begin{figure}[ht!]
    \centering
    \includegraphics[width=0.45\textwidth]{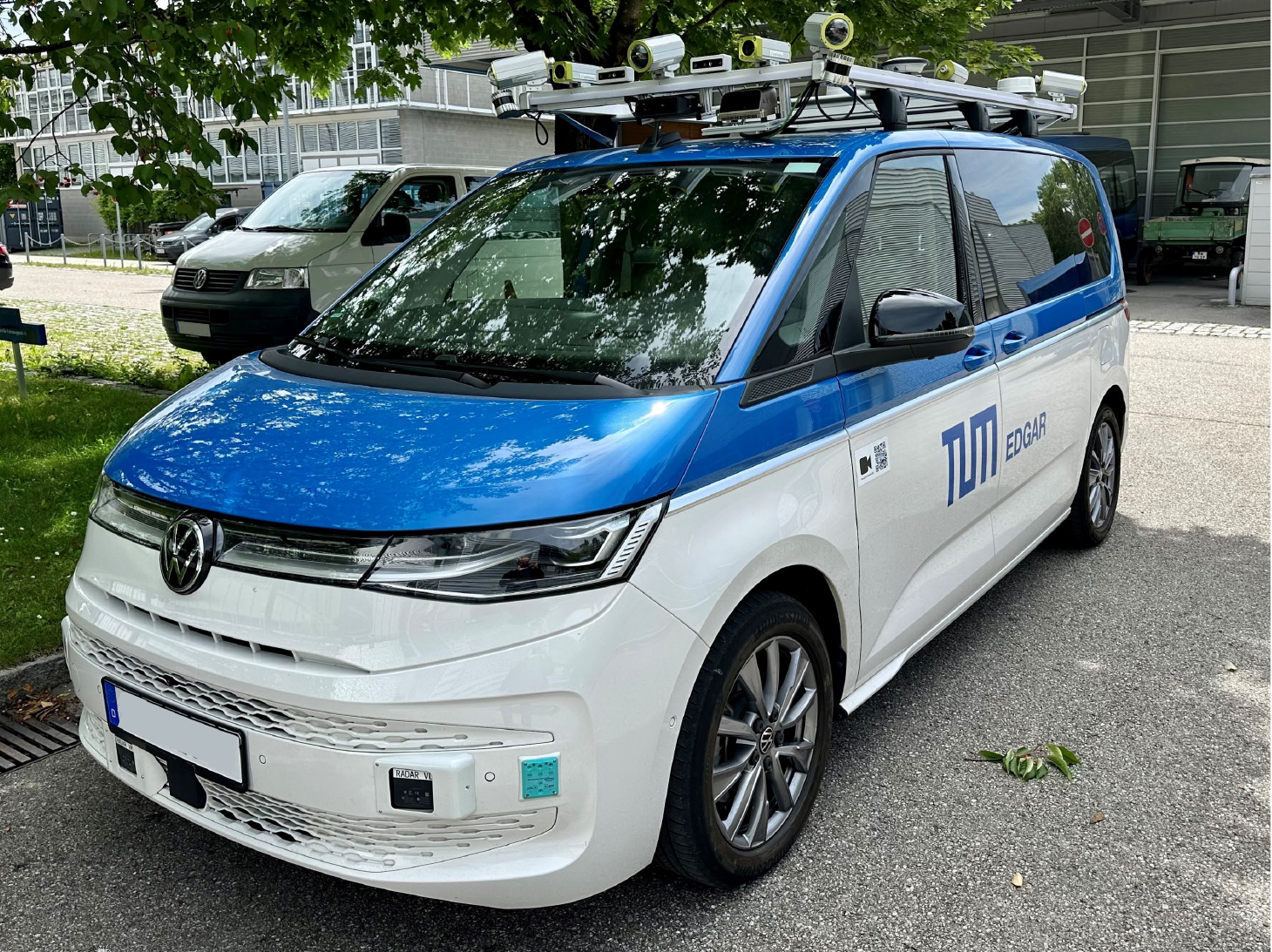}
	\caption{EDGAR research vehicle.}
	\label{fig:edgar}
\end{figure}

\begin{figure*}[ht!]
    \centering
    \includegraphics[width=0.85\textwidth]{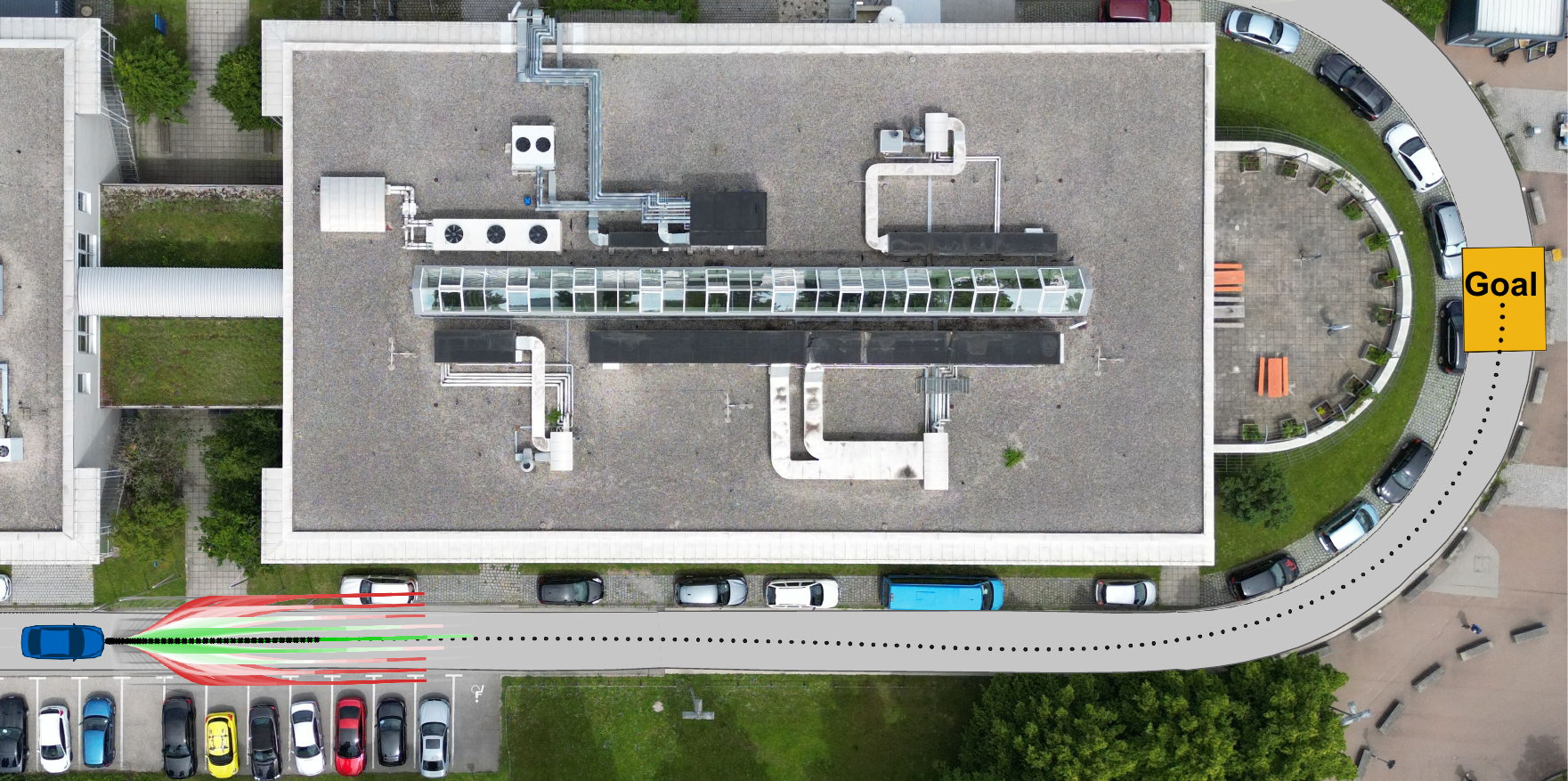}
	\caption{EDGAR research vehicle testing scenario.}
	\label{fig:route_edgar}
\end{figure*}

\begin{figure*}[ht!]
    \centering
    \resizebox{1.0\textwidth}{!}{
\begin{tikzpicture}

\definecolor{darkcyan0100189}{RGB}{0,101,189}
\definecolor{grau}{RGB}{128,128,128}
\definecolor{grün}{RGB}{162,173,0}
\definecolor{orange}{RGB}{227,114,34}
\definecolor{darkgray176}{RGB}{176,176,176}
\definecolor{black}{RGB}{0,0,0}

\begin{groupplot}[group style={group size=4 by 1, horizontal sep=1.35cm}]
\nextgroupplot[
tick align=outside,
tick pos=left,
title={$\Delta$ $v$ in $\SI{}{\meter\per\second}$},
title style={font=\huge, yshift=6pt},
xmin=0.5, xmax=1.5,
y grid style={darkgray176},
ytick={-0.05,-0.025,0,0.025,0.05,0.075,0.1},
ymin=-0.0541381472373127, ymax=0.114219186911909,
ytick style={color=black},
yticklabel style={font=\Large, /pgf/number format/fixed, /pgf/number format/precision=4},
xtick=\empty
]
\path [draw=black, fill=darkcyan0100189, thick]
(axis cs:0.75,-0.0051248789034728)
--(axis cs:1.25,-0.0051248789034728)
--(axis cs:1.25,0.0069415485412927)
--(axis cs:0.75,0.0069415485412927)
--(axis cs:0.75,-0.0051248789034728)
--cycle;
\addplot [thick, black]
table {%
1 -0.0051248789034728
1 -0.0214945095679842
};
\addplot [thick, black]
table {%
1 0.0069415485412927
1 0.0241031994123109
};
\addplot [thick, black]
table {%
0.875 -0.0214945095679842
1.125 -0.0214945095679842
};
\addplot [thick, black]
table {%
0.875 0.0241031994123109
1.125 0.0241031994123109
};
\addplot [black, mark=o, mark size=3, mark options={solid,fill opacity=0}, only marks]
table {%
1 -0.0254192992929378
1 -0.0464855411396208
1 -0.0378135197979887
1 -0.0326383705615081
1 -0.0379287100909566
1 0.0271294125229071
1 0.0269450691584485
1 0.0339703036766987
1 0.046512356745475
1 0.0585452893029244
1 0.0649171570302485
1 0.028675592251485
1 0.0321926719289418
1 0.0394538793141519
1 0.0403536564786438
1 0.031347821949185
1 0.0272065637947571
1 0.0332463495581774
1 0.0328841956181812
1 0.0323926872165754
1 0.0606110713860513
1 0.0493931452658771
1 0.106566580814217
};
\addplot [thick, black]
table {%
0.75 0.000413888537063922
1.25 0.000413888537063922
};
\addplot [black, mark=x, mark size=12.5, mark options={solid,fill=red}, only marks]
table {%
1 0.00329965934329395
};

\nextgroupplot[
tick align=outside,
tick pos=left,
title={$\Delta$ $p$ in $\SI{}{\meter}$},
title style={font=\huge, yshift=6pt},
ytick distance=0.1,
xmin=0.5, xmax=1.5,
y grid style={darkgray176},
ymin=-0.0318988826324238, ymax=0.681148891602003,
ytick style={color=black},
yticklabel style={font=\Large, /pgf/number format/fixed, /pgf/number format/precision=4},
xtick=\empty 
]
\path [draw=black, fill=grau, thick]
(axis cs:0.75,0.0400309988301547)
--(axis cs:1.25,0.0400309988301547)
--(axis cs:1.25,0.207305453609351)
--(axis cs:0.75,0.207305453609351)
--(axis cs:0.75,0.0400309988301547)
--cycle;
\addplot [thick, black]
table {%
1 0.0400309988301547
1 0.000512379832777426
};
\addplot [thick, black]
table {%
1 0.207305453609351
1 0.457993869606275
};
\addplot [thick, black]
table {%
0.875 0.000512379832777426
1.125 0.000512379832777426
};
\addplot [thick, black]
table {%
0.875 0.457993869606275
1.125 0.457993869606275
};
\addplot [black, mark=o, mark size=3, mark options={solid,fill opacity=0}, only marks]
table {%
1 0.648737629136802
};
\addplot [thick, black]
table {%
0.75 0.0773724082319229
1.25 0.0773724082319229
};
\addplot [black, mark=x, mark size=12.5, mark options={solid,fill=red}, only marks]
table {%
1 0.128245064151744
};

\nextgroupplot[
tick align=outside,
tick pos=left,
title={Longitudinal $\Delta$ $p$ in $\SI{}{\meter}$},
title style={font=\huge, yshift=6pt},
xmin=0.5, xmax=1.5,
y grid style={darkgray176},
ytick distance=0.1,
ymin=-0.685178481390655, ymax=0.126460101461863,
ytick style={color=black},
yticklabel style={font=\Large, /pgf/number format/fixed, /pgf/number format/precision=4},
xtick=\empty 
]
\path [draw=black, fill=grün, thick]
(axis cs:0.75,-0.207444023888129)
--(axis cs:1.25,-0.207444023888129)
--(axis cs:1.25,0.0755547900851639)
--(axis cs:0.75,0.0755547900851639)
--(axis cs:0.75,-0.207444023888129)
--cycle;
\addplot [thick, black]
table {%
1 -0.207444023888129
1 -0.456569473806283
};
\addplot [thick, black]
table {%
1 0.0755547900851639
1 0.0895674386049308
};
\addplot [thick, black]
table {%
0.875 -0.456569473806283
1.125 -0.456569473806283
};
\addplot [thick, black]
table {%
0.875 0.0895674386049308
1.125 0.0895674386049308
};
\addplot [black, mark=o, mark size=3, mark options={solid,fill opacity=0}, only marks]
table {%
1 -0.648285818533722
};
\addplot [thick, black]
table {%
0.75 0.0376807924496489
1.25 0.0376807924496489
};
\addplot [black, mark=x, mark size=12.5, mark options={solid,fill=red}, only marks]
table {%
1 -0.0526213788223323
};

\nextgroupplot[
tick align=outside,
tick pos=left,
title={Lateral $\Delta$ $p$ in $\SI{}{\meter}$},
title style={font=\huge, yshift=6pt},
xmin=0.5, xmax=1.5,
y grid style={darkgray176},
ymin=-0.0140096856449609, ymax=0.0152202428783087,
ytick style={color=black},
yticklabel style={font=\Large, /pgf/number format/fixed, /pgf/number format/precision=4},
xtick=\empty 
]
\path [draw=black, fill=orange, thick]
(axis cs:0.75,-0.00108875483444674)
--(axis cs:1.25,-0.00108875483444674)
--(axis cs:1.25,0.000367447442324196)
--(axis cs:0.75,0.000367447442324196)
--(axis cs:0.75,-0.00108875483444674)
--cycle;
\addplot [thick, black]
table {%
1 -0.00108875483444674
1 -0.00325511361431708
};
\addplot [thick, black]
table {%
1 0.000367447442324196
1 0.00249305050456405
};
\addplot [thick, black]
table {%
0.875 -0.00325511361431708
1.125 -0.00325511361431708
};
\addplot [thick, black]
table {%
0.875 0.00249305050456405
1.125 0.00249305050456405
};
\addplot [black, mark=o, mark size=3, mark options={solid,fill opacity=0}, only marks]
table {%
1 -0.00389099171568174
1 -0.00464445945748232
1 -0.00357739416568885
1 -0.00761676165790343
1 -0.00416239051811932
1 -0.00809878887208322
1 -0.012657691749781
1 -0.00448029782931775
1 -0.0126810525302668
1 -0.00386251584702584
1 -0.00689620645185829
1 -0.00551767598252259
1 -0.00412010931877385
1 -0.00356977041709431
1 -0.00562508278267374
1 -0.00555380126871816
1 -0.00414094795807472
1 -0.00798569128316792
1 -0.00428288292015904
1 -0.00640809496800386
1 -0.00783649312006508
1 -0.004386422672168
1 -0.00594603599891079
1 -0.00466423403826575
1 -0.00649753398410145
1 -0.00444950437922906
1 -0.00785418226673
1 -0.00787501744892061
1 0.00789217562635152
1 0.00759836714908143
1 0.00361884275479823
1 0.00362325828095088
1 0.00305814461281506
1 0.00286482579909354
1 0.0138916097636146
1 0.00387697600042917
1 0.0055660459260456
1 0.0105884739576006
};
\addplot [thick, black]
table {%
0.75 -3.35524165180723e-05
1.25 -3.35524165180723e-05
};
\addplot [black, mark=x, mark size=12.5, mark options={solid,fill=red}, only marks]
table {%
1 -0.000645910202850515
};
\end{groupplot}

\end{tikzpicture}
    }
	\caption{Deviations between the planned trajectory and the actual output of the controller. Velocity difference in \SI{}{\meter\per\second} (blue) and absolute position difference in \SI{}{\meter} (grey), divided into longitudinal (green) and lateral (orange) shown from left to right.}
	\label{fig:boxplots_edgar}
\end{figure*}
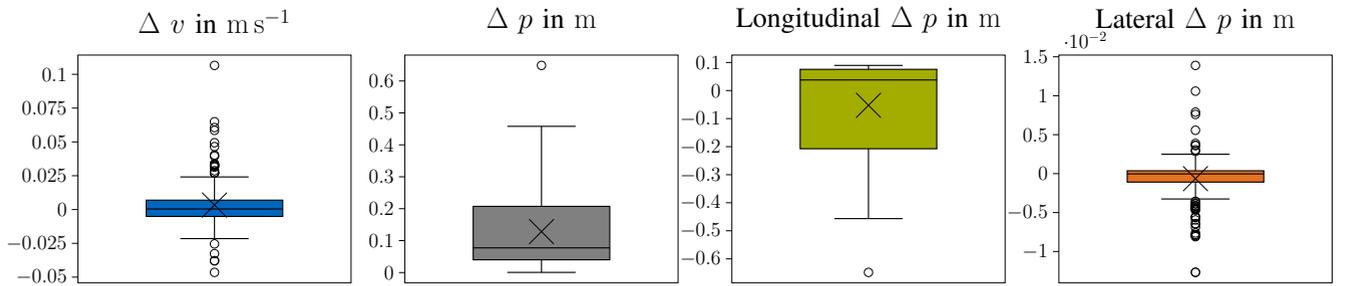

This section explores the algorithm's applicability to real-world vehicles. The tests are conducted using the EDGAR research vehicle shown in \cref{fig:edgar}. This Volkswagen T7 bus is equipped with sensors and hardware necessary for fully autonomous test runs. Detailed specifications of the vehicle are available in~\cite{karle2024edgar}.
In addition to in-house implementations, the vehicle uses the basic Autoware Universe software stack~\cite{autoware}. We use ROS2 as the middleware communication interface~\cite{Ros2}. The planning algorithm is integrated into the vehicle using a CommonRoad interface toolbox~\cite{Wuersching2024}. In this way, the planning module of the Autoware software stack is completely replaced. The planning algorithm, therefore, plans on the vehicle in the same simulation environment format~\cite{Wuersching2024}. The trajectories generated by the presented algorithm are sent to the controller via ROS2~\cite{Ros2}. The control algorithm was not fine-tuned to the research vehicle and the trajectory planning algorithm to demonstrate adaptivity~\cite{autoware}. 
To analyze the algorithm, we initiate acceleration from a standstill, drive into a curve, and then apply braking while within the curve. \cref{fig:route_edgar} shows the driven route. The yellow area represents the destination region, and the blue vehicle shows the starting position. 
\cref{fig:route_edgar} presents a combination of a real-world drone image and the CommonRoad lanelet street format. The algorithm utilizes the CommonRoad scenario format to plan the next trajectory~\cite{commonroad,Wuersching2024}. \cref{fig:route_edgar} illustrates the initial planning step of the vehicle along with the sampled trajectories. Gray represents dynamically infeasible trajectories, while the gradient from red to green indicates the varying costs of the feasible trajectories. The black path signifies the selected trajectory for this time step.
\cref{fig:boxplots_edgar} illustrates the discrepancies between the planned trajectory, as proposed by our algorithms, and the actual output from the vehicle's controller in autonomous driving scenarios.
It can be seen that the velocity and position differences are small at moderate acceleration. The difference in velocity is usually less than \SI{0.025}{\meter\per\second}.
The position deviations in the lateral and longitudinal directions differ. There is a greater deviation from the controller in the longitudinal direction than in the lateral direction.
The update frequency of the overall software and the driving direction can explain individual deviation measurements. The longitudinal positioning may exhibit deviations from the intended trajectory, attributable to the frequency at which the trajectory is replanned. Enhancing the frequency of trajectory planning enables more rapid integration of any deviations by the control system into the updated trajectory for the vehicle. Consequently, increasing the planning frequency can effectively diminish these deviations.

\section{Discussion}
\label{sec:discussion}
\subsection{FRENETIX Performance}
The results from our study indicate that the FRENETIX framework presented is effective in rapidly finding a safe and reliable trajectory in dynamic environments. By outsourcing the computationally intensive functionalities to C++, the performance can be kept high even with many objects, so sufficient trajectories can always be generated at any time. However, it is essential to note that specific settings and parameters influence the algorithm's performance, which is consistent with expectations for an analytical algorithm. During our research, collisions and other issues were mainly attributed to the lack of features for specific scenarios or inadequate parameter tuning. Nonetheless, the potential for optimizing vehicle behavior in various situations is achievable through further extensions and modules. The cost functions demonstrate a variable influence on the target variables, facilitating the successful simulation of complex maneuvers, such as overtaking in the presence of oncoming traffic. 

\subsection{FRENETIX Applicability} 
The presented open-source toolbox demonstrates its capability to effectively perform tests on research vehicles. Its modular structure facilitates the rapid adaptation and integration of new features and modules. This toolbox provides a convenient platform for testing and validation, addressing individual planning challenges. The accelerated research enabled by this work focuses on a modular software stack for real-world autonomous driving investigations between perception and control modules as well as for pure simulation purposes. Beyond simulation, the algorithm has been successfully implemented in a real research vehicle, with the sampling method efficiently identifying feasible trajectories. The results confirm that the controller can follow the planner's trajectory. However, it is noted that the resulting deviations can be minimized further, as the controller was not specifically tuned to the vehicle or the planning algorithm. It should also be noted that the system's overall behavior depends on many factors and is difficult to attribute to the planning algorithm. Furthermore, the presented work could also be used to investigate algorithms on other vehicles like small-scale RC cars or ground robots. 
\section{Conclusion \& Outlook}
\label{sec:conclusion}
This paper introduced FRENETIX, a high-performance and modular sampling-based trajectory planner algorithm designed for autonomous driving applications. We propose a novel integration of multiple steps and methods to develop a trajectory planner that operates with high computational efficiency. FRENETIX is characterized by its robustness, adaptability, and capacity to handle complex scenarios through individualized extensions. Its modular design and the presented cost functions facilitate varied prioritizations of driving behavior by adjusting the cost weights and validity checks.
Our experimental results demonstrate the algorithm's ability to produce dynamic vehicle behaviors, such as overtaking maneuvers, particularly in highly dynamic environments, including scenarios with oncoming traffic. By using FRENETIX on a research vehicle, we demonstrate its suitability for use outside of simulation purposes. 
By open-sourcing FRENETIX, we aim to provide a valuable motion planning baseline for the community, fostering collaborative development and innovation.
FRENETIX lays a strong foundation for future research in areas such as behavior planning, reinforcement learning, and other extensions, offering significant potential for comprehensive benchmark analyses. Further investigations could explore the algorithm's performance in diverse and more complex traffic situations, potentially leading to enhancements in autonomous driving systems.

\printbibliography[]

@article{Dijkstra.1959,
	doi = {10.1007/bf01386390},
	year = {1959},
	month = dec,
	publisher = {Springer Science and Business Media {LLC}},
	volume = {1},
	number = {1},
	pages = {269--271},
	author = {E. W. Dijkstra},
	title = {A note on two problems in connexion with graphs},
	journal = {Numerische Mathematik}
}

@article{Hart.1968,
	author={Hart, Peter E. and Nilsson, Nils J. and Raphael, Bertram},
	journal={IEEE Transactions on Systems Science and Cybernetics}, 
	title={A Formal Basis for the Heuristic Determination of Minimum Cost Paths}, 
	year={1968},
	volume={4},
	number={2},
	pages={100-107},
	doi={10.1109/TSSC.1968.300136}
}

@inproceedings{Naumann.2020,
	author={Naumann, Maximilian and Sun, Liting and Zhan, Wei and Tomizuka, Masayoshi},
	booktitle={2020 IEEE International Conference on Robotics and Automation (ICRA)}, 
	title={Analyzing the Suitability of Cost Functions for Explaining and Imitating Human Driving Behavior based on Inverse Reinforcement Learning}, 
	year={2020},
	pages={5481-5487},
	doi={10.1109/ICRA40945.2020.9196795}
}

@INPROCEEDINGS{Takahashi,
  author={Takahashi, A. and Hongo, T. and Ninomiya, Y. and Sugimoto, G.},
  booktitle={Proceedings. IEEE/RSJ International Workshop on Intelligent Robots and Systems '. (IROS '89) 'The Autonomous Mobile Robots and Its Applications}, 
  title={Local Path Planning And Motion Control For Agv In Positioning}, 
  year={1989},
  volume={},
  number={},
  pages={392-397},
  doi={10.1109/IROS.1989.637936}
}

@article{Trauth2023,
  title = {Toward Safer Autonomous Vehicles: Occlusion-Aware Trajectory Planning to Minimize Risky Behavior},
  volume = {4},
  ISSN = {2687-7813},
  DOI = {10.1109/ojits.2023.3336464},
  journal = {IEEE Open Journal of Intelligent Transportation Systems},
  publisher = {Institute of Electrical and Electronics Engineers (IEEE)},
  author = {Trauth,  Rainer and Moller,  Korbinian and Betz,  Johannes},
  year = {2023},
  pages = {929–942}
}

@INPROCEEDINGS{DrivabilityChecker,
	author={Pek, Christian and Rusinov, Vitaliy and Manzinger, Stefanie and Can Üste, Murat and Althoff, Matthias},
	title={CommonRoad Drivability Checker: Simplifying the Development and Validation of Motion Planning Algorithms},
	pages={1-8},
	booktitle={Proc. of the IEEE Intelligent Vehicles Symposium},
	year={2020},
}

@INPROCEEDINGS{commonroad,
  author={Althoff, Matthias and Koschi, Markus and Manzinger, Stefanie},
  booktitle={2017 IEEE Intelligent Vehicles Symposium (IV)}, 
  title={CommonRoad: Composable benchmarks for motion planning on roads}, 
  year={2017},
  volume={},
  number={},
  pages={719-726},
  doi={10.1109/IVS.2017.7995802  }
}

@inproceedings{walenet,
	author = {Geisslinger, Maximilian and Karle, Phillip and Betz, Johannes and Lienkamp, Markus},
	title = {Watch-and-Learn-Net: Self-supervised Online Learning for Probabilistic Vehicle Trajectory Prediction},
	booktitle = {2021 IEEE International Conference on Systems, Man, and Cybernetics (SMC)},
	year = {2021},
	publisher = {IEEE},
	doi = {10.1109/smc52423.2021.9659079 },
}

@article{geisslingerconcept,
	author = {Geisslinger, Maximilian and Poszler, Franziska and Betz, Johannes and Lütge, Christoph and Lienkamp, Markus},
	title = {Autonomous Driving Ethics: from Trolley Problem to Ethics of Risk},
	journal = {Philosophy and Technology},
	year = {2021},
	doi = {10.1007/s13347-021-00449-4},
}

@article{Elallid2022,
    title = {A Comprehensive Survey on the Application of Deep and Reinforcement Learning Approaches in Autonomous Driving},
    journal = {Journal of King Saud University - Computer and Information Sciences},
    volume = {34},
    number = {9},
    pages = {7366-7390},
    year = {2022},
    issn = {1319-1578},
    doi = {https://doi.org/10.1016/j.jksuci.2022.03.013},
    author = {Badr Ben Elallid and Nabil Benamar and Abdelhakim Senhaji Hafid and Tajjeeddine Rachidi and Nabil Mrani},
}

@ARTICLE{Aradi2022,
  author={Aradi, Szilárd},
  journal={IEEE Transactions on Intelligent Transportation Systems}, 
  title={Survey of Deep Reinforcement Learning for Motion Planning of Autonomous Vehicles}, 
  year={2022},
  volume={23},
  number={2},
  pages={740-759},
  doi={10.1109/TITS.2020.3024655}}

@INPROCEEDINGS{Trauth-RL,
      title={A Reinforcement Learning-Boosted Motion Planning Framework: Comprehensive Generalization Performance in Autonomous Driving}, 
      booktitle={2024 IEEE Intelligent Vehicles Symposium (IV)}, 
      author={Rainer Trauth and Alexander Hobmeier and Johannes Betz},
      year={2024},
      pages = {1--8},
      organization = {IEEE},
      note = {Accepted, preprint available: \url{https://arxiv.org/abs/2402.01465}}
}

@INPROCEEDINGS{kaufeld2024investigating,
      title={Investigating Driving Interactions: A Robust Multi-Agent Simulation Framework for Autonomous Vehicles}, 
      booktitle={2024 IEEE Intelligent Vehicles Symposium (IV)}, 
      author={Marc Kaufeld and Rainer Trauth and Johannes Betz},
      year={2024},
      pages = {1--8},
      organization = {IEEE},
      note = {Accepted, preprint available: \url{https://arxiv.org/abs/2402.04720}}
}

@article{Zhou2022,
  title = {A review of motion planning algorithms for intelligent robots},
  author = {Zhou, Chengmin and Huang, Bingding and Fr{\"a}nti, Pasi},
  journal = {Journal of Intelligent Manufacturing},
  volume = {33},
  number = {2},
  pages = {387--424},
  year = {2022},
  issn = {1572-8145},
  date = {2022-02-01},
  id = {Zhou2022}

}

@ARTICLE{Dong2023,
  author={Dong, Lu and He, Zichen and Song, Chunwei and Sun, Changyin},
  journal={Journal of Systems Engineering and Electronics}, 
  title={A review of mobile robot motion planning methods: from classical motion planning workflows to reinforcement learning-based architectures}, 
  year={2023},
  volume={34},
  number={2},
  pages={439-459},
  doi={10.23919/JSEE.2023.000051}
}

@Article{Rowold2022,
AUTHOR = {Rowold, Matthias and Ögretmen, Levent and Kerbl, Tobias and Lohmann, Boris},
TITLE = {Efficient Spatiotemporal Graph Search for Local Trajectory Planning on Oval Race Tracks},
JOURNAL = {Actuators},
VOLUME = {11},
YEAR = {2022},
NUMBER = {11},
ARTICLE-NUMBER = {319},
ISSN = {2076-0825},
DOI = {10.3390/act11110319}
}

@article{werling.2012,
author = {Moritz Werling and Sören Kammel and Julius Ziegler and Lutz Gröll},
title ={Optimal trajectories for time-critical street scenarios using discretized terminal manifolds},
journal = {The International Journal of Robotics Research},
volume = {31},
number = {3},
issn = {0278-3649},
pages = {346-359},
year = {2012},
doi = {10.1177/0278364911423042},
}

@INPROCEEDINGS{werling.2010,
  author={Werling, Moritz and Ziegler, Julius and Kammel, Sören and Thrun, Sebastian},
  booktitle={2010 IEEE International Conference on Robotics and Automation}, 
  title={Optimal trajectory generation for dynamic street scenarios in a Frenét Frame}, 
  year={2010},
  volume={}, 
  isbn = {978-1-4244-5038-1},
  publisher = {IEEE},
  number={},
  pages={987-993},
  doi={10.1109/ROBOT.2010.5509799}
}

@ARTICLE{Huang2023,
  author={Huang, Jianyu and He, Zuguang and Arakawa, Yutaka and Dawton, Billy},
  journal={IEEE Access}, 
  title={Trajectory Planning in Frenet Frame via Multi-Objective Optimization}, 
  year={2023},
  volume={11},
  number={},
  pages={70764-70777},
  doi={10.1109/ACCESS.2023.3294713}
}

@ARTICLE{Li2023,
  author={Li, Yanfeng},
  journal={IEEE Access}, 
  title={Motion Planning for Dynamic Scenario Vehicles in Autonomous-Driving Simulations}, 
  year={2023},
  volume={11},
  number={},
  pages={2035-2047},
  doi={10.1109/ACCESS.2022.3233822}
}

@ARTICLE{Teng2023,
  author={Teng, Siyu and et al.},
  journal={IEEE Transactions on Intelligent Vehicles}, 
  title={Motion Planning for Autonomous Driving: The State of the Art and Future Perspectives}, 
  year={2023},
  volume={8},
  number={6},
  pages={3692-3711},
  doi={10.1109/TIV.2023.3274536}
}

@Article{Albarella2023,
AUTHOR = {Albarella, Nicola and Lui, Dario Giuseppe and Petrillo, Alberto and Santini, Stefania},
TITLE = {A Hybrid Deep Reinforcement Learning and Optimal Control Architecture for Autonomous Highway Driving},
JOURNAL = {Energies},
VOLUME = {16},
YEAR = {2023},
NUMBER = {8},
ARTICLE-NUMBER = {3490},
ISSN = {1996-1073},
DOI = {10.3390/en16083490}
}

@ARTICLE{Gonzlez2015,
author={González, David and Pérez, Joshué and Milanés, Vicente and Nashashibi, Fawzi},
journal={IEEE Transactions on Intelligent Transportation Systems},
title={A Review of Motion Planning Techniques for Automated Vehicles},
year={2016},
volume={17},
number={4},
pages={1135-1145},
doi={10.1109/TITS.2015.2498841}
}

@article{Pokorny2022,
  title = {Descriptive analysis of reports on autonomous vehicle collisions in California: January 2021–June 2022},
  volume = {2},
  ISSN = {2004-3082},
  DOI = {10.55329/xydm4000},
  journal = {Traffic Safety Research},
  publisher = {Dept. of Technology & Society,  Faculty of Engineering,  LTH,  Lund University},
  author = {Pokorny,  Petr and Høye,  Alena},
  year = {2022},
  month = {9},
  pages = {000011}
}

@INPROCEEDINGS{Gu2013,
  author={Gu, Tianyu and Snider, Jarrod and Dolan, John M. and Lee, Jin-woo},
  booktitle={2013 IEEE Intelligent Vehicles Symposium (IV)}, 
  title={Focused Trajectory Planning for autonomous on-road driving}, 
  year={2013},
  volume={},
  number={},
  pages={547-552},
  doi={10.1109/IVS.2013.6629524}}

@article{Katrakazas2015,
title = {Real-time motion planning methods for autonomous on-road driving: State-of-the-art and future research directions},
journal = {Transportation Research Part C: Emerging Technologies},
volume = {60},
pages = {416-442},
year = {2015},
issn = {0968-090X},
doi = {https://doi.org/10.1016/j.trc.2015.09.011}  ,
author = {Christos Katrakazas and Mohammed Quddus and Wen-Hua Chen and Lipika Deka},
}

@INPROCEEDINGS{Stahl2019,
  author={Stahl, Tim and Wischnewski, Alexander and Betz, Johannes and Lienkamp, Markus},
  booktitle={2019 IEEE Intelligent Transportation Systems Conference (ITSC)}, 
  title={Multilayer Graph-Based Trajectory Planning for Race Vehicles in Dynamic Scenarios}, 
  year={2019},
  volume={},
  number={},
  pages={3149-3154},
  doi={10.1109/ITSC.2019.8917032}}

@ARTICLE{Liang2015,
  author={Ma, Liang and Xue, Jianru and Kawabata, Kuniaki and Zhu, Jihua and Ma, Chao and Zheng, Nanning},
  journal={IEEE Transactions on Intelligent Transportation Systems}, 
  title={Efficient Sampling-Based Motion Planning for On-Road Autonomous Driving}, 
  year={2015},
  volume={16},
  number={4},
  pages={1961-1976},
  doi={10.1109/TITS.2015.2389215}}

@article{Chen2019,
  title = {Parallel planning: a new motion planning framework for autonomous driving},
  volume = {6},
  ISSN = {2329-9274},
  DOI = {10.1109/jas.2018.7511186},
  number = {1},
  journal = {IEEE/CAA Journal of Automatica Sinica},
  publisher = {Institute of Electrical and Electronics Engineers (IEEE)},
  author = {Chen,  Long and Hu,  Xuemin and Tian,  Wei and Wang,  Hong and Cao,  Dongpu and Wang,  Fei-Yue},
  year = {2019},
  month = jan,
  pages = {236–246}
}

@INPROCEEDINGS{Schoemer2020,
  author={Schömer, Philip and Hüneberg, Mark Timon and Zöllner, J. Marius},
  booktitle={2020 IEEE Intelligent Vehicles Symposium (IV)}, 
  title={Optimization of Sampling-Based Motion Planning in Dynamic Environments Using Neural Networks}, 
  year={2020},
  volume={},
  number={},
  pages={2110-2117},
  doi={10.1109/IV47402.2020.9304644}}

@ARTICLE{Li2022,
  author={Li, Bai and Acarman, Tankut and Zhang, Youmin and Ouyang, Yakun and Yaman, Cagdas and Kong, Qi and Zhong, Xiang and Peng, Xiaoyan},
  journal={IEEE Transactions on Intelligent Transportation Systems}, 
  title={Optimization-Based Trajectory Planning for Autonomous Parking With Irregularly Placed Obstacles: A Lightweight Iterative Framework}, 
  year={2022},
  volume={23},
  number={8},
  pages={11970-11981},
  doi={10.1109/TITS.2021.3109011}}

@INPROCEEDINGS{Ziegler20214,
  author={Ziegler, Julius and Bender, Philipp and Dang, Thao and Stiller, Christoph},
  booktitle={2014 IEEE Intelligent Vehicles Symposium Proceedings}, 
  title={Trajectory planning for Bertha — A local, continuous method}, 
  year={2014},
  volume={},
  number={},
  pages={450-457},
  doi={10.1109/IVS.2014.6856581}}

@inproceedings{Wuersching2024,
	author = {Würsching, Gerald and Mascetta, Tobias and Lin, Yuanfei and Althoff, Matthias},
	title = {Simplifying Sim-to-Real Transfer in Autonomous Driving: Coupling Autoware with the CommonRoad Motion Planning Framework},
	booktitle = {2024 IEEE Intelligent Vehicles Symposium (IV)},
	year = {2024},
}

@misc{karle2024edgar,
      title={EDGAR: An Autonomous Driving Research Platform -- From Feature Development to Real-World Application}, 
      author={Phillip Karle and Tobias Betz and Marcin Bosk and Felix Fent and Nils Gehrke and Maximilian Geisslinger and Luis Gressenbuch and Philipp Hafemann and Sebastian Huber and Maximilian Hübner and Sebastian Huch and Gemb Kaljavesi and Tobias Kerbl and Dominik Kulmer and Tobias Mascetta and Sebastian Maierhofer and Florian Pfab and Filip Rezabek and Esteban Rivera and Simon Sagmeister and Leander Seidlitz and Florian Sauerbeck and Ilir Tahiraj and Rainer Trauth and Nico Uhlemann and Gerald Würsching and Baha Zarrouki and Matthias Althoff and Johannes Betz and Klaus Bengler and Georg Carle and Frank Diermeyer and Jörg Ott and Markus Lienkamp},
      year={2024},
      archivePrefix={arXiv},
      url={https://arxiv.org/pdf/2309.15492}
}

@article{Ros2,
    author = {Steven Macenski and Tully Foote and Brian Gerkey and Chris Lalancette and William Woodall},
    title = {Robot Operating System 2: Design, architecture, and uses in the wild},
    journal = {Science Robotics},
    volume = {7},
    number = {66},
    pages = {eabm6074},
    year = {2022},
    doi = {10.1126/scirobotics.abm6074}
}

@INPROCEEDINGS{autoware,
  author={Kato, Shinpei and Tokunaga, Shota and Maruyama, Yuya and Maeda, Seiya and Hirabayashi, Manato and Kitsukawa, Yuki and Monrroy, Abraham and Ando, Tomohito and Fujii, Yusuke and Azumi, Takuya},
  booktitle={2018 ACM/IEEE 9th International Conference on Cyber-Physical Systems (ICCPS)}, 
  title={Autoware on Board: Enabling Autonomous Vehicles with Embedded Systems}, 
  year={2018},
  volume={},
  number={},
  pages={287-296},
  doi={10.1109/ICCPS.2018.00035}
}

@article{dierckx1982algorithms,
  title={Algorithms for smoothing data with periodic and parametric splines},
  author={Dierckx, P.},
  journal={Computer Graphics and Image Processing},
  volume={20},
  pages={171--184},
  year={1982}
}

@techreport{dierckx1981algorithms,
  title={Algorithms for smoothing data with periodic and parametric splines},
  author={Dierckx, P.},
  year={1981},
  institution={Dept. Computer Science, K.U.Leuven},
  number={tw55}
}

@book{dierckx1993curve,
  title={Curve and surface fitting with splines},
  author={Dierckx, P.},
  series={Monographs on Numerical Analysis},
  year={1993},
  publisher={Oxford University Press}
}

@ARTICLE{Manzinger2021,
  author={Manzinger, Stefanie and Pek, Christian and Althoff, Matthias},
  journal={IEEE Transactions on Intelligent Vehicles}, 
  title={Using Reachable Sets for Trajectory Planning of Automated Vehicles}, 
  year={2021},
  volume={6},
  number={2},
  pages={232-248},
  doi={10.1109/TIV.2020.3017342}
}

@report{Paden.2016,
 author = {Paden, Brian and Cap, Michal and Yong, Sze Zheng and Yershov, Dmitry and Frazzoli, Emilio},
 year = {2016},
 month = {4},
 day = {25},
 title = {A Survey of Motion Planning and Control Techniques for Self-driving  Urban Vehicles},
 url = {http://arxiv.org/pdf/1604.07446v1}
}

@article{GARCIA1989,
    title = {Model predictive control: Theory and practice—A survey},
    journal = {Automatica},
    volume = {25},
    number = {3},
    pages = {335-348},
    year = {1989},
    issn = {0005-1098},
    doi = {https://doi.org/10.1016/0005-1098(89)90002-2},
    author = {Carlos E. García and David M. Prett and Manfred Morari},
}

@inproceedings{Klimke2023,
  author={Klimke, Marvin and Völz, Benjamin and Buchholz, Michael},
  booktitle={2023 IEEE Intelligent Vehicles Symposium (IV)}, 
  title={Integration of Reinforcement Learning Based Behavior Planning With Sampling Based Motion Planning for Automated Driving}, 
  year={2023},
  volume={},
  number={},
  pages={1-8}
}

@misc{bojarski2016end,
      title={End to End Learning for Self-Driving Cars}, 
      author={Mariusz Bojarski and Davide Del Testa and Daniel Dworakowski and Bernhard Firner and Beat Flepp and Prasoon Goyal and Lawrence D. Jackel and Mathew Monfort and Urs Muller and Jiakai Zhang and Xin Zhang and Jake Zhao and Karol Zieba},
      year={2016},
      eprint={1604.07316},
      archivePrefix={arXiv},
      primaryClass={cs.CV}
}

@article{Xin2021,
    author = {Long Xin and Yiting Kong and Shengbo Eben Li and Jianyu Chen and Yang Guan and Masayoshi Tomizuka and Bo Cheng},
    title ={Enable faster and smoother spatio-temporal trajectory planning for autonomous vehicles in constrained dynamic environment},
    journal = {Proceedings of the Institution of Mechanical Engineers, Part D: Journal of Automobile Engineering},
    volume = {235},
    number = {4},
    pages = {1101-1112},
    year = {2021},
    doi = {10.1177/0954407020906627},
}

@ARTICLE{Muhammad2021,
  author={Muhammad, Khan and Ullah, Amin and Lloret, Jaime and Ser, Javier Del and de Albuquerque, Victor Hugo C.},
  journal={IEEE Transactions on Intelligent Transportation Systems}, 
  title={Deep Learning for Safe Autonomous Driving: Current Challenges and Future Directions}, 
  year={2021},
  volume={22},
  number={7},
  pages={4316-4336},
  doi={10.1109/TITS.2020.3032227}
}

@INPROCEEDINGS{Ichter2018,
  author={Ichter, Brian and Harrison, James and Pavone, Marco},
  booktitle={2018 IEEE International Conference on Robotics and Automation (ICRA)}, 
  title={Learning Sampling Distributions for Robot Motion Planning}, 
  year={2018},
  volume={},
  number={},
  pages={7087-7094},
  doi={10.1109/ICRA.2018.8460730}
}

@INPROCEEDINGS{Huang2017,
  author={Huang, Eric and Mukadam, Mustafa and Liu, Zhen and Boots, Byron},
  booktitle={2017 IEEE International Conference on Robotics and Automation (ICRA)}, 
  title={Motion planning with graph-based trajectories and Gaussian process inference}, 
  year={2017},
  volume={},
  number={},
  pages={5591-5598},
  doi={10.1109/ICRA.2017.7989659}
}

@Inbook{Gasparetto2015,
  author    = {Alessandro Gasparetto and Paolo Boscariol and Albano Lanzutti and Renato Vidoni},
  editor    = {Giuseppe Carbone and Fernando Gomez-Bravo},
  title     = {Path Planning and Trajectory Planning Algorithms: A General Overview},
  bookTitle = {Motion and Operation Planning of Robotic Systems: Background and Practical Approaches},
  year      = {2015},
  publisher = {Springer International Publishing},
  address   = {Cham},
  pages     = {3--27},
  isbn      = {978-3-319-14705-5},
  doi       = {10.1007/978-3-319-14705-5_1},
}

@ARTICLE{Tampuu2022,
  author={Tampuu, Ardi and Matiisen, Tambet and Semikin, Maksym and Fishman, Dmytro and Muhammad, Naveed},
  journal={IEEE Transactions on Neural Networks and Learning Systems}, 
  title={A Survey of End-to-End Driving: Architectures and Training Methods}, 
  year={2022},
  volume={33},
  number={4},
  pages={1364-1384},
  doi={10.1109/TNNLS.2020.3043505}
}


\vspace{-2cm}

\begin{IEEEbiography}[{\includegraphics[width=1in,height=1.25in,clip,keepaspectratio]{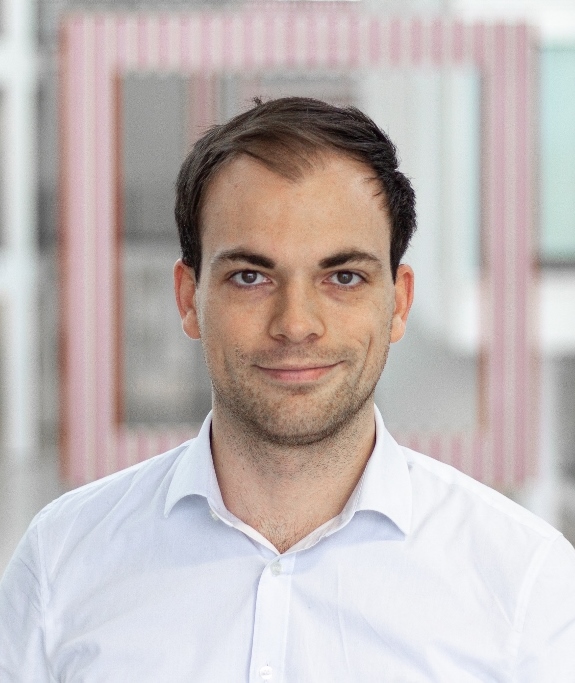}}]{\textbf{Rainer Trauth}}~received a B.Sc. degree in engineering science and an M.Sc. in mechanical engineering from the Technical University of Munich (TUM) in 2017 and 2020, respectively, where he is currently pursuing a Ph.D. degree in mechanical engineering at the Institute of Automotive Technology.
His research interests include motion planning, situational awareness, and behavior planning approaches focusing on real-world applications in autonomous driving.
\end{IEEEbiography}

\vspace{-2cm}

\begin{IEEEbiography}[{\includegraphics[width=1in,height=1.25in,clip,keepaspectratio]{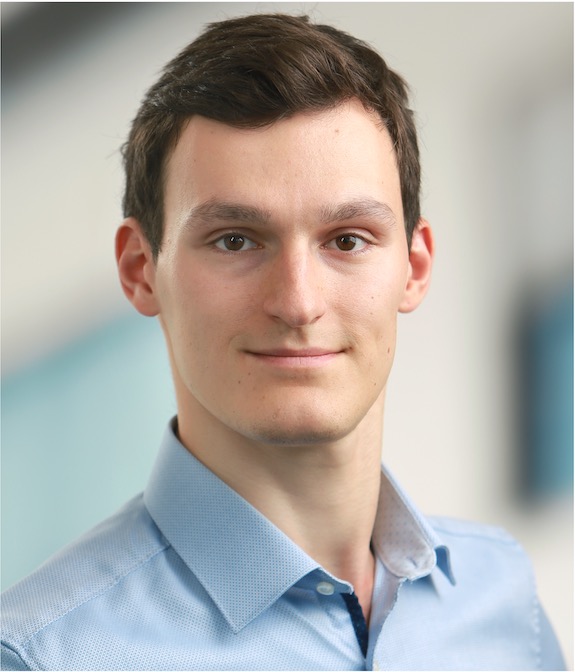}}]{\textbf{Korbinian Moller}}
~received a B.Sc. degree in mechanical engineering from the Technical University of Munich (TUM) in 2021 and a M.Sc. in 2023 degree in mechanical engineering at the TUM School of Engineering and Design. He is currently pursuing a Ph.D. degree as part of the Autonomous Vehicle Systems (AVS) lab at TUM.
His research interests include edge-case scenario simulation, the optimization of vehicle behavior, and motion planning in autonomous driving.
\end{IEEEbiography}

\vspace{-2cm}

\begin{IEEEbiography}[{\includegraphics[width=1in,height=1.25in,clip,keepaspectratio]{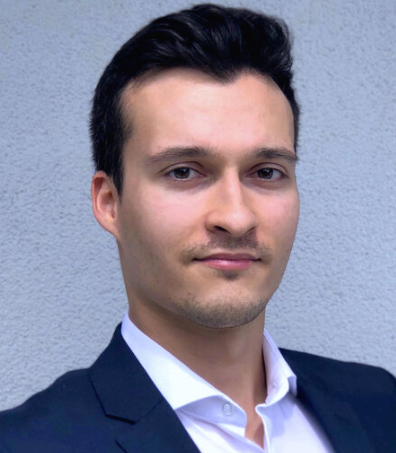}}]{\textbf{Gerald Würsching}}
received a B.Sc. degree and an M.Sc. degree in mechanical engineering from the Technical University of Munich (TUM), Munich, Germany, in 2018 and 2020, respectively. He is currently pursuing a Ph.D. degree at the professorship of cyber-physical systems (CPS) at TUM. His research interests include motion planning for autonomous driving, particularly developing robust and efficient planning algorithms for arbitrary traffic situations.
\end{IEEEbiography}

\newpage

\begin{IEEEbiography}[{\includegraphics[width=1in,height=1.25in,clip,keepaspectratio]{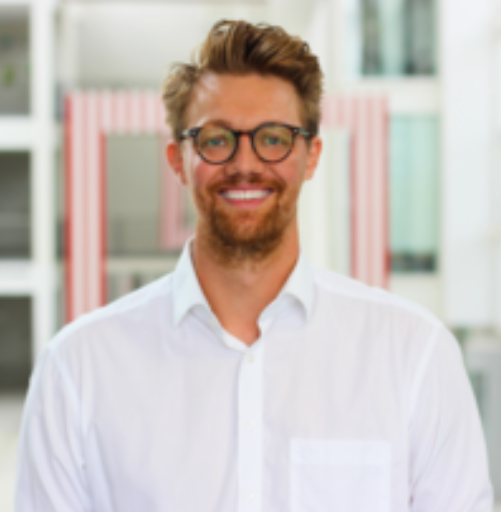}}]{\textbf{Johannes Betz}}
~is an assistant professor in the Department of Mobility Systems Engineering at the Technical University of Munich (TUM), leading the Autonomous Vehicle Systems (AVS) lab. He is one of the founders of the TUM Autonomous Motorsport team. His research focuses on developing adaptive dynamic path planning and control algorithms, decision-making algorithms that work under high uncertainty in multi-agent environments, and validating the algorithms on real-world robotic systems. Johannes earned a B.Eng. (2011) from the University of Applied Science Coburg, an M.Sc. (2012) from the University of Bayreuth, an MA (2021) in philosophy from TUM, and a Ph.D. (2019) from TUM. 
\end{IEEEbiography}


\EOD

\end{document}